\documentclass{article}

\usepackage{arxiv}

\usepackage[utf8]{inputenc} 
\usepackage[T1]{fontenc}    
\usepackage{hyperref}       
\usepackage{url}            
\usepackage{booktabs}       
\usepackage{amsfonts}       
\usepackage{nicefrac}       
\usepackage{microtype}      
\usepackage{hyperref}
\hypersetup{
    colorlinks=true,
    linkcolor=blue,
    filecolor=magenta,      
    urlcolor=red,
}
\usepackage{float}
\usepackage{dblfloatfix}
\usepackage{verbatim}
\usepackage{pgfplots}
\pgfplotsset{compat=1.14} 
\usepackage{setspace}
\usepackage{array}
\usepackage{amsmath,amssymb,amsfonts}

\usepackage[caption=false, font=footnotesize]{subfig}

\title{A Hybrid Approach and Unified Framework for Bibliographic Reference Extraction}

\author{
  Syed Tahseen Raza Rizvi\\
  German Research Center for Artificial Intelligence (DFKI)\\
  Kaiserslautern, Germany \\
  \texttt{syed\_tahseen\_raza.rizvi@dfki.de} \\
   \And
 Andreas Dengel \\
  German Research Center for Artificial Intelligence (DFKI)\\
  Kaiserslautern, Germany \\
  \texttt{andreas.dengel@dfki.de} \\
   \And
 Sheraz Ahmed \\
  German Research Center for Artificial Intelligence (DFKI)\\
  Kaiserslautern, Germany \\
  \texttt{sheraz.ahmed@dfki.de} \\
}

\begin{document}
\maketitle

\begin{abstract}
Publications are an integral part in a scientific community. Bibliographic reference extraction from scientific publication is a challenging task due to diversity in referencing styles and document layout. Existing methods perform sufficiently on one dataset however, applying these solutions to a different dataset proves to be challenging. Therefore, a generic solution was anticipated which could overcome the limitations of the previous approaches. The contribution of this paper is three-fold. First, it presents a novel approach called \textit{DeepBiRD} which is inspired by human visual perception and exploits layout features to identify individual references in a scientific publication. Second, we release a large dataset for image-based reference detection with $2401$ scans containing $38863$ references, all manually annotated for individual reference. Third, we present a unified and highly configurable end-to-end automatic bibliographic reference extraction framework called \textit{BRExSys} which employs \textit{DeepBiRD} along with state-of-the-art text-based models to detect and visualize references from a bibliographic document. Our proposed approach pre-processes the images in which a hybrid representation is obtained by processing the given image using different computer vision techniques. Then, it performs layout driven reference detection using Mask R-CNN on a given scientific publication. \textit{DeepBiRD} was evaluated on two different datasets to demonstrate the generalization of this approach. The proposed system achieved an AP50 of $98.56\%$ on our dataset. \textit{DeepBiRD} significantly outperformed the current state-of-the-art approach on their dataset. Therefore, suggesting that \textit{DeepBiRD} is significantly superior in performance, generalized, and independent of any domain or referencing style.
\end{abstract}


\begin{figure*}[!tbh]
\centering
\subfloat[Text-based approach (ParsCit) on APA style]{\includegraphics[width=0.45\linewidth]{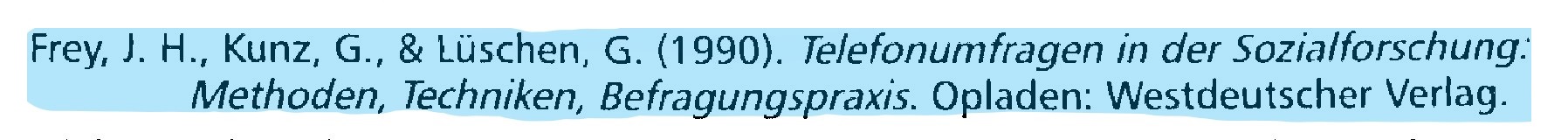}\label{fig:parscit_on_apa}}\hfil
\subfloat[Layout-based (Proposed) approach on APA style]{\includegraphics[width=0.45\linewidth]{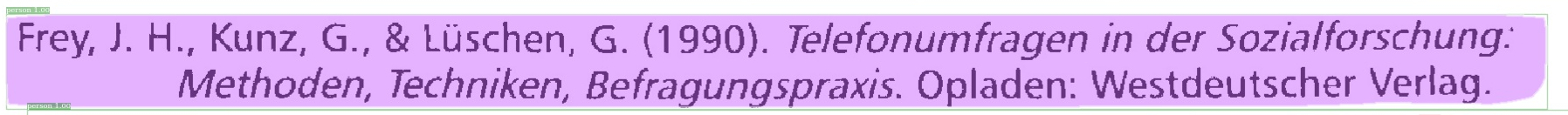}\label{fig:deepbird_on_apa}}\hfil
\\
\subfloat[Text-based approach (ParsCit) on Hybrid style]{\includegraphics[width=0.45\linewidth]{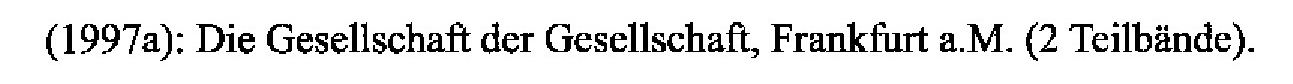}\label{fig:parscit_on_hybrid}}\hfil
\subfloat[Layout-based (Proposed) approach on Hybrid style]{\includegraphics[width=0.45\linewidth]{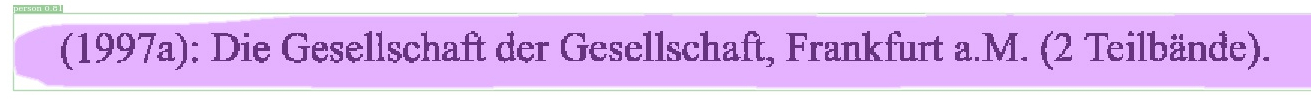}\label{fig:deepbird_on_hybrid}}\hfil
\\
\subfloat[Text-based approach (ParsCit) on Alpha style]{\includegraphics[width=0.45\linewidth]{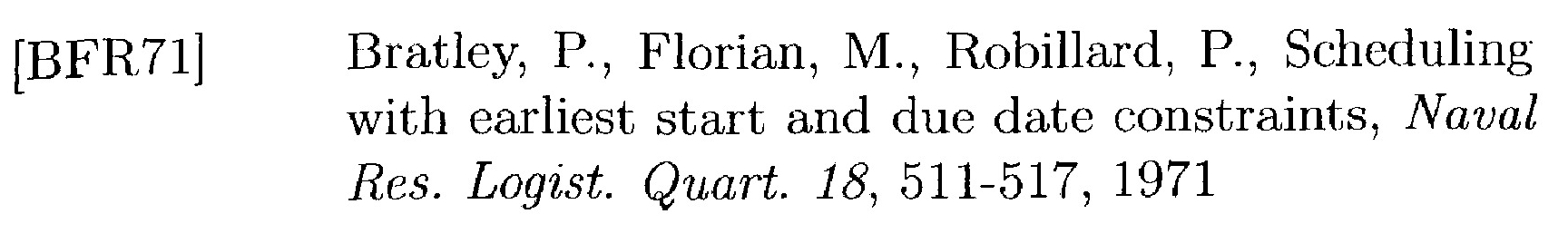}\label{fig:parscit_on_alpha}}\hfil
\subfloat[Layout-based (Proposed) approach on Alpha style]{\includegraphics[width=0.45\linewidth]{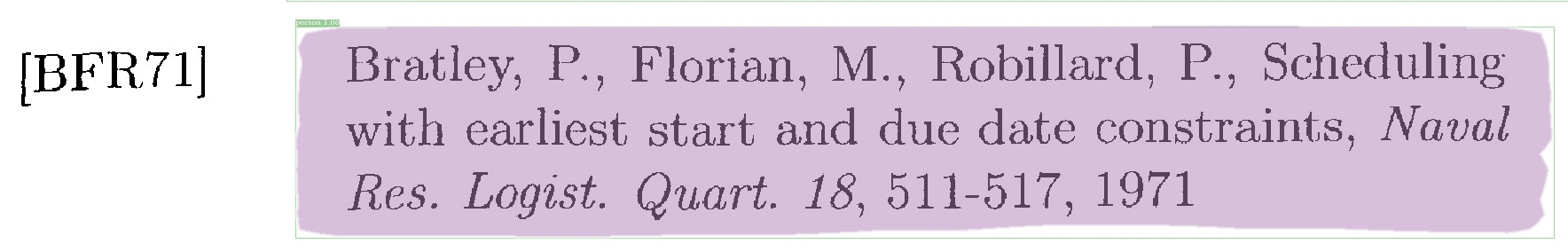}\label{fig:deepbird_on_alpha}}\hfil
\hfil
\caption{Detection results of text-based and layout-based methods}
\label{fig:text_vs_image_based_solutions}
\end{figure*}

\section{Introduction}
\label{sec:introduction}
There has been a rapid increase in every field of research since the start of the 21st century, subsequently increasing the volume of scientific literature exponentially \cite{stats1,stats2}. Each scientific publication consists of several components i.e. header, abstract, sections, and references. Bibliographic references play a vital role in every publication as they serve as a citation network and forms a foundation of the information provided in the publication.

Bibliographic references are of particular interest to library communities \cite{Lauscher:2018:LOC:3197026.3197050}. They play a key role in compiling library catalogs. These catalogs contain information regarding all bibliographic items like books, journals, conference proceedings, magazines, and other media present in a library. For such purpose, it is not feasible to manually find and index such a huge volume of references. 

Resource Discovery Systems pose to be a viable solution for the libraries to further expanding their horizon by providing the indexed data available from external resources \cite{Lauscher:2018:LOC:3197026.3197050}.  Some resources are commercial and are thus paid to use their collected data i.e. Web of Science, Scopus. According to a scientometric study \cite{Mongeon2016}, both Scopus and Web of Science have mostly coverage of English journal articles from the Biomedical and Social Science domains and thus have low overall coverage for journals and articles from other domains and languages. Therefore making the Resource Discovery Systems a sub-optimal solution for bibliographic cataloging.

The majority of the related work on the problem of bibliographic reference detection are text-based solutions and therefore make use of textual features like author names, publication titles, etc. in a document to detect references. Text-based approaches use a set of carefully crafted heuristics and regular expressions \cite{citation-parser,10.1007/978-3-642-33290-6_40}  based on the position of constituents of a reference string i.e. author names, affiliations, publisher, journal/book/conference name, year of publishing, etc. 
These heuristics and regular expressions are very sensitive to the textual features as they do not anticipate any number or special character like brackets etc at the start of a reference string thus such references are missed altogether. Text-based approaches are also very sensitive to the layout of the references for example if all lines of reference strings start from the same point then text-based approaches find it very hard to identify the starting and ending boundaries of a reference string thus either detecting multiple references as one or identifying one reference as multiple references. With the introduction of such cases, carefully crafted heuristics become deprecated right away thus making text-based approaches less robust and eventually not generalizable. For instance, most common referencing styles like MLA and APA have author names and publication titles as the starting features of a reference string. On the other hand, there exist some rare bibliography styles in social sciences such as Alpha or hybrid Chicago in which reference strings either start with a reference identifier or publication year respectively. It is an example of a problematic case for text-based approaches because the references in the given example do not comply with the other common traditional referencing styles and are rarely used. Comparison of results from text-based and layout-based approaches on different referencing styles is shown in Fig \ref{fig:text_vs_image_based_solutions}. It can be observed that the text-based approach was unable to detect an unusual reference string as it entirely relies on textual features while overlooking other important facets i.e. layout features.

This paper introduces an automatic, effective, and generalized approach for reference detection from document images. It works equally well for scanned and digital-born PDFs documents. Our approach is inspired by the way how human beings perceive and identify objects. To understand this phenomenon, consider an example of an illegible blurred document containing some text. Although the document is unreadable, however we can still identify paragraphs, bullet points and similarly references. The underlying idea states that layout information is the key to identify different textual structures in a document even without using textual features. For this task, we employed Convolutional Neural Networks (CNN) for representation learning based on the layout of a document. This waives the dependency on textual heuristics used in the majority of the existing systems. Our approach is generic and is thus applicable to any bibliographic publication independent of its domain or referencing style. We also release a benchmark dataset for bibliographic reference detection from document images. We performed benchmark tasks to evaluate the performance of \textit{DeepBiRD} from different aspects i.e. generalization, robustness, etc.

In this paper, we also present a comprehensive framework called BRExSys, which encapsulates all state-of-the-art bibliographic reference detection methods under a single umbrella, allowing users to use any of the existing or proposed methods at one place. BRExSys supports scientific publications in a number of file formats i.e. born-digital PDF, Scanned PDF/images, HTML, XML etc. BRExSys provides a user-friendly interface to facilitate the smooth processing of the input file and visualization of processed output. BRExSys is a highly customizable system as it can be tailored based on the user's requirement.

The contributions of this publication are as follows:
\begin{itemize}
    \item We present a novel layout driven approach for automatic reference detection from scientific publications, which effectively exploits the visual cues to firstly identify bibliographic references from a given scientific publication.
    \item We release a new \& larger dataset for image-based reference detection which will be publicly available for the community.
    \item We demonstrate the superiority of the proposed approach by carrying out a series of comparative performance evaluations against the previous state-of-the-art approach.
    \item We present an automatic bibliographic reference detection framework called BRExSys which has integrated \textit{DeepBiRD} and other state-of-the-art text-based reference detection models to take advantage of different modalities for the task in hand.
\end{itemize}

The rest of the paper is structured as follows: Section 2 discusses the relevant work done so far on the problem of reference detection. Section 3 discusses the details of the datasets used in this work. Section 4 presents the architecture and pipeline of our proposed approach. Section 5 discusses different experimental setups used for this publication along with analyses of the results obtained from different experiments to demonstrate the effectiveness of our approach. Section 6 discusses the details of our proposed bibliographic reference extraction system called \textit{BRExSys}. And lastly, Section 7 makes the concluding remarks of this paper.

\begin{figure*}
\centering
\subfloat[Single Column Scan]{\frame{\includegraphics[height=7.2cm,width=0.31\linewidth]{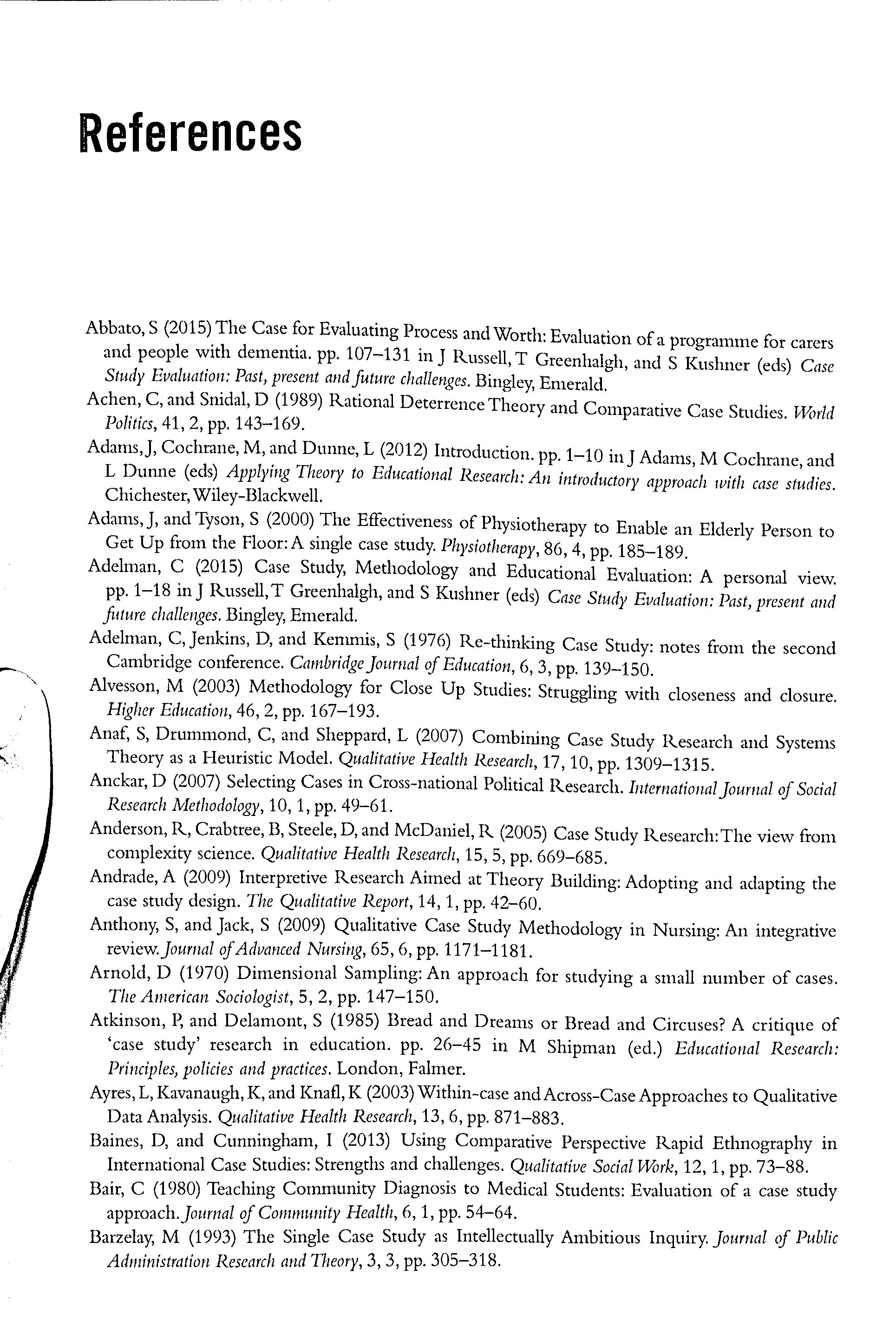}}\label{fig:single_column_scan}}\hfil
\subfloat[Double Column Scan]{\frame{\includegraphics[height=7.2cm,width=0.31\linewidth]{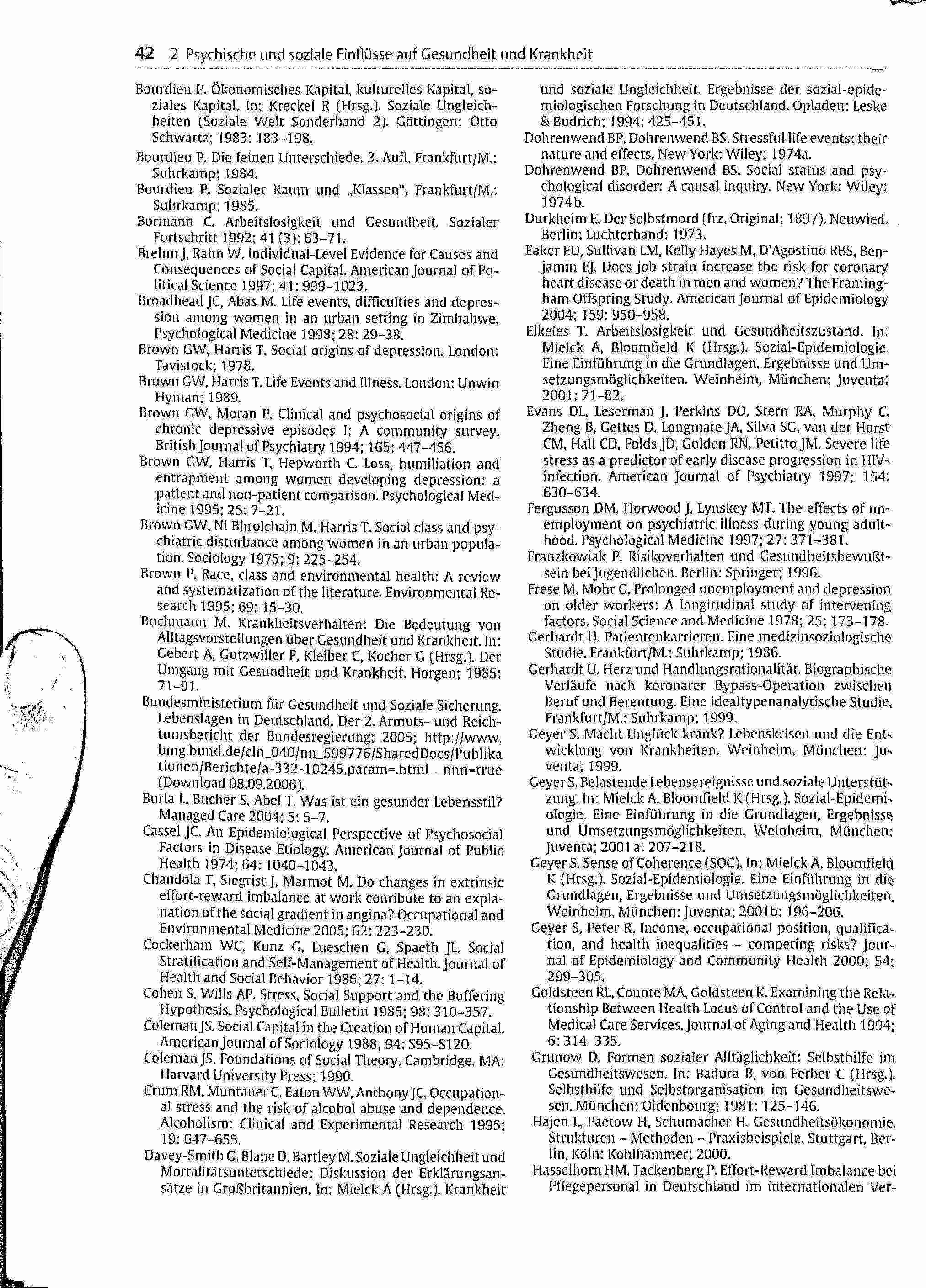}}\label{fig:double_column_scan}}\hfil
\subfloat[Triple Column Scan]{\frame{\includegraphics[height=7.2cm,width=0.31\linewidth]{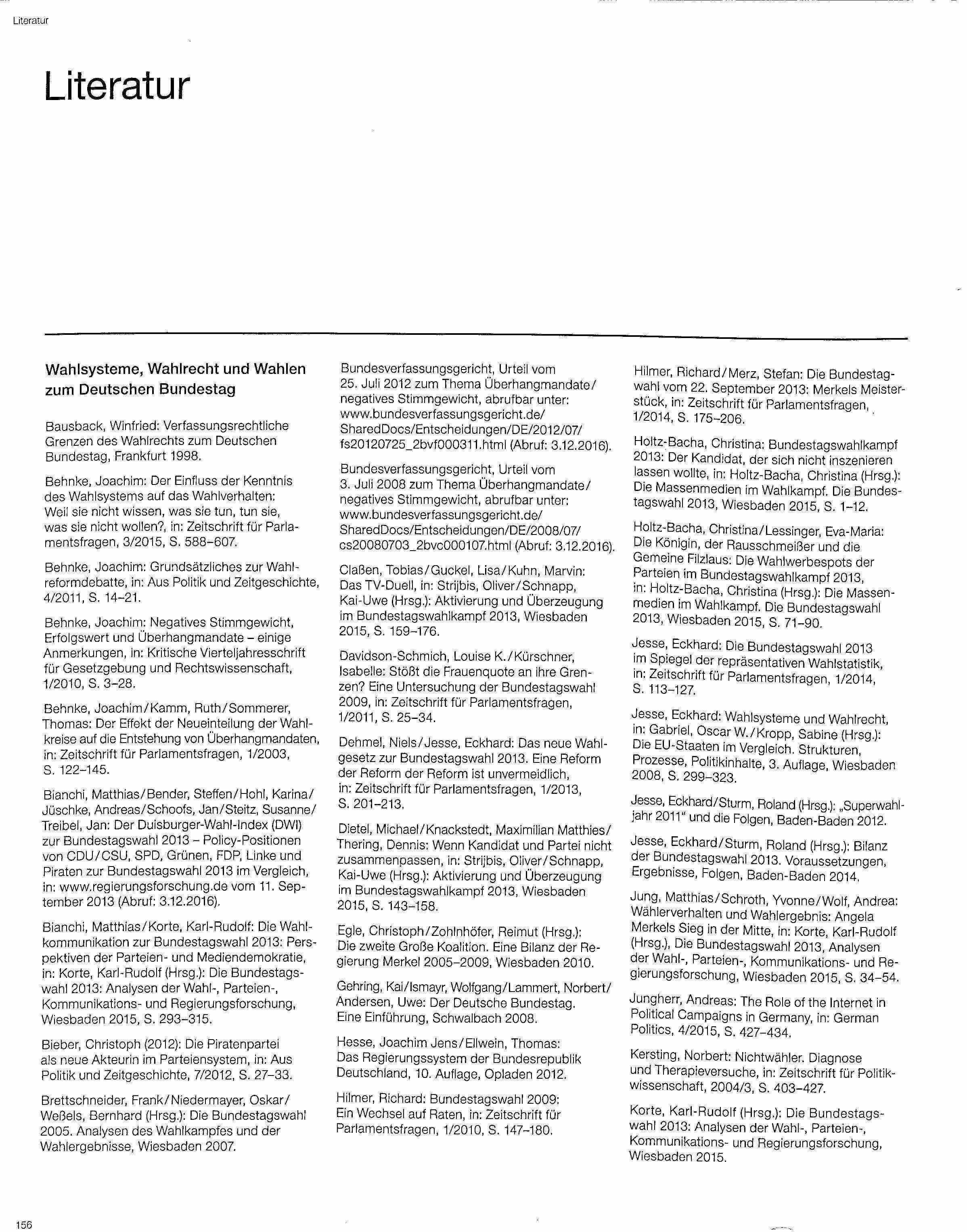}}\label{fig:triple_column_scan}}\hfil
\caption{Samples of different layouts from input files}
\label{fig:input_samples}
\end{figure*}

\section{Related Work}
\label{related-work}
A lot of work has been done in the field of reference detection. Bibliographic reference detection is generally performed by two methods like text-based and layout-based. Most of the approaches are based on the analysis of textual content to identify references. There are several techniques employed by each text-based approach to identify references. Here we will discuss such techniques used for bibliographic reference detection, starting from simplest and moving towards more sophisticated ones.

\subsection{Text-based Approaches}
The simplest of the text-based reference detection techniques employ regular expressions and carefully crafted heuristics \cite{citation-parser} for this task. Such approaches are mostly not considered as an optimal solution because of their limited coverage. For example, \textit{MLA} and \textit{APA} are the most common referencing styles in which a reference string starts with author names. To detect such references, adopted heuristics will look for comma-separated author names at the start of the reference string. The drawback of such an approach is that it will be unable to detect a reference if it does not comply with the defined heuristics i.e. reference string with \textit{Alpha} style where reference string starts with a custom ID. Every domain has its unique referencing style and sometimes there are multiple referencing styles within one domain. Such challenges make simple approaches unsuitable for this complex task.

\textit{Citation-Parser}~\cite{citation-parser} is a typical example of heuristics based tool. To identify components of bibliographic reference string i.e. authors, title, conference/journal, etc, it employs a set of carefully designed heuristics. Sautter \textit{et al.} \cite{10.1007/978-3-642-33290-6_40} proposed a tool named \textit{RefParse} which exploits similarities between individual reference strings to identify different referencing style for parsing a reference string. Perl also provided an extension named \textit{Biblio} \cite{biblio} for parsing and extracting reference string metadata. Chen \textit{et al.}~\cite{5645620} proposed \textit{BibPro}, an approach that identifies citation style by matching it with referencing styles available in its database and then uses gene sequence alignment technique to identify components of reference strings. \textit{AnyStyle-Parser} \cite{anystyle} is another example of a tool which identifies bibliographic references using heuristics. \textit{PDFSSA4MET} \cite{PDFSSA4MET} proposed a slightly different approach to identify references in a Born-digital PDF. In this approach, textual PDF is firstly converted into an XML file. Then by employing pattern matching mechanisms, syntactic and structural analysis of XML is performed to identify the reference section. 

Ahmed \textit{et al.}~\cite{9102282} used a diverse range of features i.e. font type, neighbor distance, text location, font typography and lexical properties to identify components of a scientific publication and later extract metadata like Authors name, affiliation, email, headings etc.  Boukhers \textit{et al.}~\cite{8791225} proposed an approach which all text lines are individually classified using pre-trained random forest model with the probability to be a potential reference line and later uses format, lexical, semantic and shape features to identify and segment reference strings.

Lafferty \textit{et al.} \cite{Lafferty:2001:CRF:645530.655813} proposed an advanced approach known as Conditional Random Fields (CRF). CRF is a probabilistic approach for labeling sequence data like reference strings. This labeling includes identifying different parts of a reference string i.e. authors, publication title, year, conference/journal name, etc. Such labeling assists in recognizing a reference string based on its labeled components.

Tkaczyk \textit{et al.} \cite{Tkaczyk2015} proposed "Content ExtRactor and MINEr (\textit{CERMINE})" a CRF based system for extracting and mining bibliographic metadata from references in born-digital PDF scientific articles. Free-cite \cite{freecite} computes features from tokenized citation string and then classify that token sequence using trained CRF.
Science Parse \cite{scienceparse} is a tool based on CRF to identify and extract metadata of references from a document.
Matsuoka \textit{et al.} \cite{7829774} demonstrated the use of lexical features by CRF results to gain an increase in accuracy.
Councill \textit{et al.} \cite{CouncillGK08} presented a CRF based package called "\textit{ParsCit}" for reference metadata tagging problem. In which the reference strings were identified from plain text, based on fine-grained heuristics. \cite{CouncillGK08} claims \textit{ParsCit} to be one of the best known and widely used open-source system based on Heuristics and CRF for reference detection, string parsing, and metadata tagging. Tkaczyk \textit{et al.} \cite{Tkaczyk2018ParsRecAN} also proposed a reference metadata recommender system which provided 10 most popular open-source citation parser tools in one system. Selected tools were a mixture of simple heuristics based and machine learning-based solutions.

Nowadays, artificial neural networks are the most popular choice as a solution to most of the scientific problems. Similarly, some literature also explored the potential of neural networks for the task of bibliographic reference detection and parsing.

Zou \textit{et al.} \cite{Zou2010} proposed a two steps approach to locate and parse bibliographic references in HTML medical articles. In the first step individual references are located using machine learning approaches whereas in the second step by employing CRFs, metadata is extracted from each reference.

Contrary to the traditional approaches for reference tagging, Parsad \textit{et al.} \cite{animesh2018neuralparscit} proposed a bibliographic reference string parser named "\textit{Neural-ParsCit}" based on deep neural networks. The authors tried to capture long-range dependencies in reference strings using Long Short Term Memory (LSTM) \cite{Hochreiter:1997:LSM:1246443.1246450} based architecture. Lopez \textit{et al.} \cite{10.1007/978-3-642-04346-8_62} proposed a tool named "\textit{Grobid}" based a tool based on conditional random fields for detection and extraction of publication headers, bibliographic references and their respective metadata. The \textit{Grobid} model was trained on multi-domain, manually annotated data containing $6835$ instances. Recently, Grennan \textit{et al.}~\cite{grennan2020synthetic} performed experiments to train a CRF-based solution\mbox{~\cite{10.1007/978-3-642-04346-8_62}} on actual citation parsing data annotated by humans and synthetic data and suggested that the model trained on synthetic dataset performed very similar to the model trained on original data.

Text-based approaches are not directly applicable to document images. To identify references from scanned documents, the text must be extracted from a given document by performing Optical Character Recognition (OCR) and then applying the selected approach to extracted text. The disadvantage of this approach is the potential introduction of OCR error which will eventually contribute to detection error thus making the task unnecessarily complicated.

\subsection{Layout-based Approaches}
The literature discussed so far relies only on textual features to identify references. Text-based approaches do not take advantage of layout features thus abandoning an important aspect. There are very few approaches that explored the potential of exploiting layout information for detecting bibliographic references.

\begin{table}
\caption{Overall distribution of \textit{BibX}\cite{ICONIP} dataset}
\centering
\label{bibx_classification2}
\begin{tabular}{llll}
\hline\noalign{\smallskip}
 & Train Set & Validation Set & Test Set  \\
\noalign{\smallskip}\hline\noalign{\smallskip}
No. of Images & 287 & 25 & 143 \\
No. of References & 5741 & 478 & 2547 \\
\noalign{\smallskip}\hline\noalign{\smallskip}
Single Column & 270 & 24 & 136 \\
Double Column & 17 & 1 & 7 \\
\noalign{\smallskip}\hline
\end{tabular}
\end{table}

Bhardwaj \textit{et al.} \cite{10.1007/978-3-319-70096-0_30} used layout information to detect references from a scanned document. For that purpose, Fully Convolutional Neural Network (FCN) \cite{7298965} was used to segment the references and later post-processed to identify individual references. To our best knowledge, it is currently the state-of-the-art for the image-based reference detection task. The authors also released a small dataset  \cite{ICONIP} for image-based reference detection. In this paper, this dataset will be referred to as \textit{BibX} dataset. Lauscher \textit{et al.} \cite{Lauscher:2018:LOC:3197026.3197050} used this layout based reference detection in their system \cite{locdb} to build an open database of citations for libraries indexing use case. Recently, Rizvi \textit{et al.} \cite{dicta19Ref} gauged the performance of four state-of-the-art object detection models using layout information to detect bibliographic references in a scientific publication.

\section{Datasets}
\label{dataset}
\subsection{BibX dataset}
\label{bibx}
This section provides insights about the dataset used for training and baseline performance comparison of \textit{DeepBIBX} \cite{10.1007/978-3-319-70096-0_30} and our proposed approach \textit{DeepBiRD} for the task of layout-based reference detection. To the best of the authors' knowledge, it is the only image-based dataset that contains annotations of references. \textit{BibX} dataset consists of 455 document images from several social sciences books and journals, containing 429 and 25 document image samples from single and double column layouts respectively. The dataset is divided into train, validation, and test set with 287, 25, and 143 samples respectively. Distribution details of \textit{BibX} dataset are mentioned in Table \ref{bibx_classification2}. Furthermore, considering the limited size of the dataset, we propose a new dataset called \textit{BibLy} dataset. Details of this new dataset are discussed in the following section.

\begin{table}[!b]
\caption{Overall distribution of \textit{BibLy} dataset}
\centering
\label{tab:2}
\begin{tabular}{llll}
\hline\noalign{\smallskip}
 & Train Set & Validation Set & Test Set  \\
\noalign{\smallskip}\hline\noalign{\smallskip}
No. of Images & 1513 & 132 & 756 \\
No. of References & 24606 & 2013 & 12244 \\
\noalign{\smallskip}\hline\noalign{\smallskip}
Single Column & 1411 & 124 & 705 \\
Double Column & 92 & 7 & 46 \\
Triple Column & 10 & 1 & 5 \\
\noalign{\smallskip}\hline
\end{tabular}
\end{table}

\subsection{BibLy dataset}
\label{refdet}
In this paper, we are releasing a dataset named \textit{BibLy} \cite{NeuezugangDB} for image-based reference detection. This dataset has been curated from the reference section of various Journals, Monographs, Articles, and Books from the social sciences domain. The resolution of images varies from 1500 to 4500 for the larger side of the image. Image quality is maintained on at least 300 dpi. All images were manually annotated where a box was drawn around every single reference. 

There are 2,401 scanned document images in \textit{BibLy} dataset containing 38,863 references in total. Document scans were initially divided into three groups based on the number of columns i.e. single, double, and triple columns. Table \ref{tab:2} shows the distribution of samples in layout groups. These groups were further distributed into train, validation, and test set with balanced representation from each group. Fig \ref{fig:input_samples} shows sample scans with different layouts. 
The distribution of train test and validation set along with their respective number of references are shown in Table \ref{tab:2}. Dataset is shared on the following link: \url{https://madata.bib.uni-mannheim.de/283/}.

\begin{figure*}[!t]
\centering
  \includegraphics[width=0.9\textwidth]{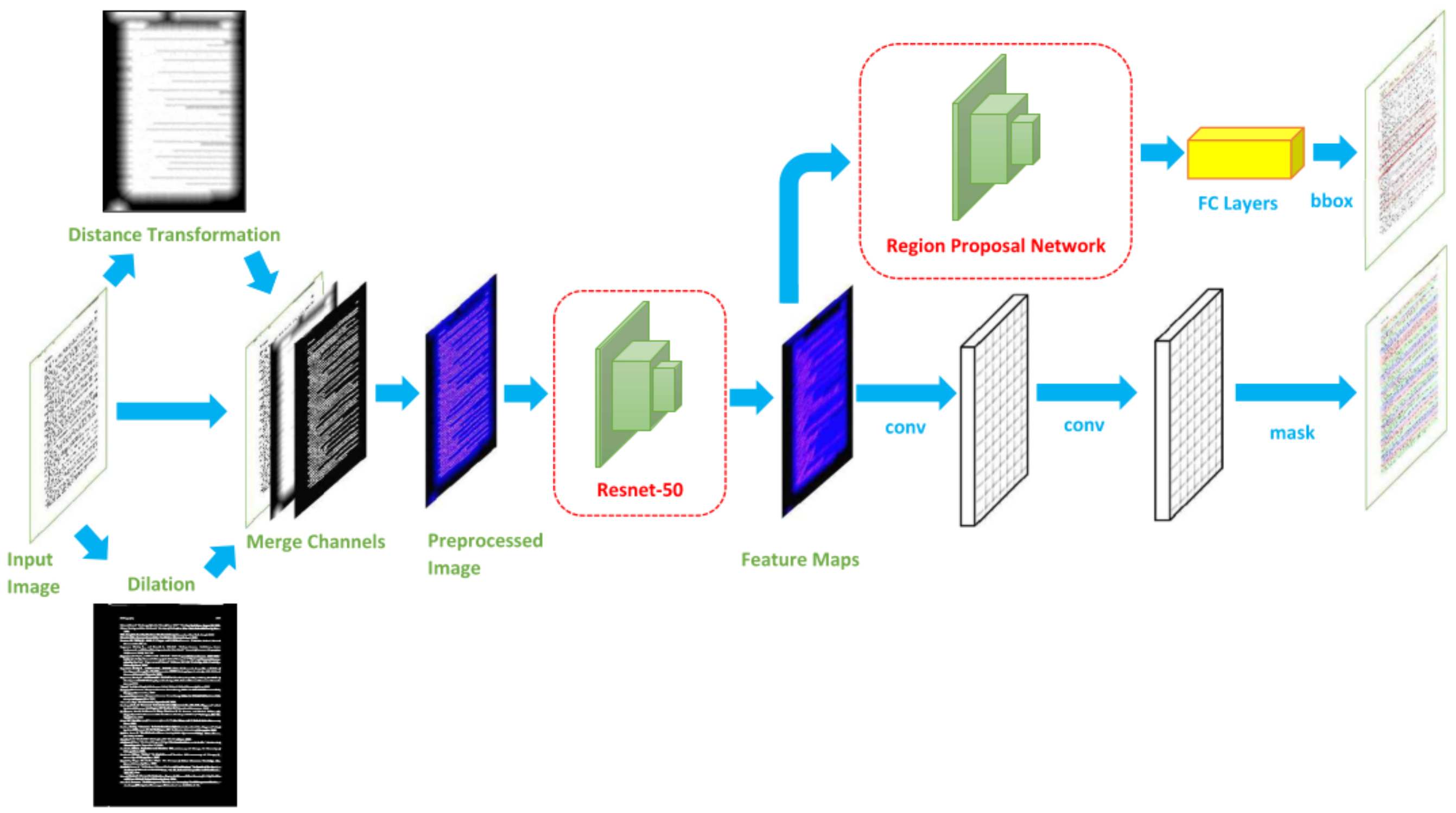}
\caption{Overview of proposed \textit{DeepBiRD} pipeline for layout-based reference detection}
\label{fig:8}
\end{figure*}

\section{DeepBiRD: Proposed Approach}
\label{proposed-approach}
In our proposed approach, we exploit layout information to detect references from a given document image. Firstly, we pre-process the input document image and incorporate more layout information to facilitate bibliographic reference detection. Later, references are detected from each pre-processed image. Fig \ref{fig:8} depicts the complete pipeline of our proposed system. Details of our pipeline are discussed as follows.

\subsection{Pre-processing}
\label{sec:preprocessing}
The first stage in our pipeline is pre-processing followed by the reference detection. In order to highlight layout features, we obtained a hybrid representation, which highlights the important content and helps the automatic representation learning approach to extract discriminative features. This hybrid representation is achieved by applying different transformations on the input image. The pre-processing stage involves a series of steps, which are elaborated as follows:

\subsubsection{Distance Transform}
\label{sec:distancetransform}

Distance transform provides the distance between each pixel and the nearest input foreground pixel. This way, we can highlight the separation between words, lines, and characters which later proves to help identify and separate individual references.

To apply distance transform, we firstly read the image as a grayscale image followed by inversion of the image therefore switching bright pixels with dark pixels and vice versa. Then we binarize the input image using OTSU thresholding followed by inversion of all pixel values. Distance transform is then applied to the resultant binarized inverted image. We used several distance types in different experiments and selected Euclidean distance with a $3\times3$ mask as the most suitable distance measure. An example of euclidean distance transformation is shown in Fig \ref{fig:dist-transform-example}.

\subsubsection{Dilation}
\label{sec:dilation}
We performed dilation on the input image to highlight text regions along with their surroundings to facilitate the neural network to identify lines and their respective scope more precisely. To perform dilation, we firstly binarized the input image using OTSU thresholding followed by inversion. Then we perform dilation using a kernel of $1\times5$, this horizontal kernel merges the nearby characters in the proximity of the same line. The motivation of using a kernel of $1\times5$ is to preserve line separation while merging the words in the same line, therefore highlighting a line. A sample image is shown in Fig \ref{fig:dilated-example}.

\subsubsection{Hybrid Representation}
\label{sec:mergechannels}
\begin{figure*}
\centering
\subfloat[Dilated]{\frame{\includegraphics[height=7cm,width=0.32\linewidth]{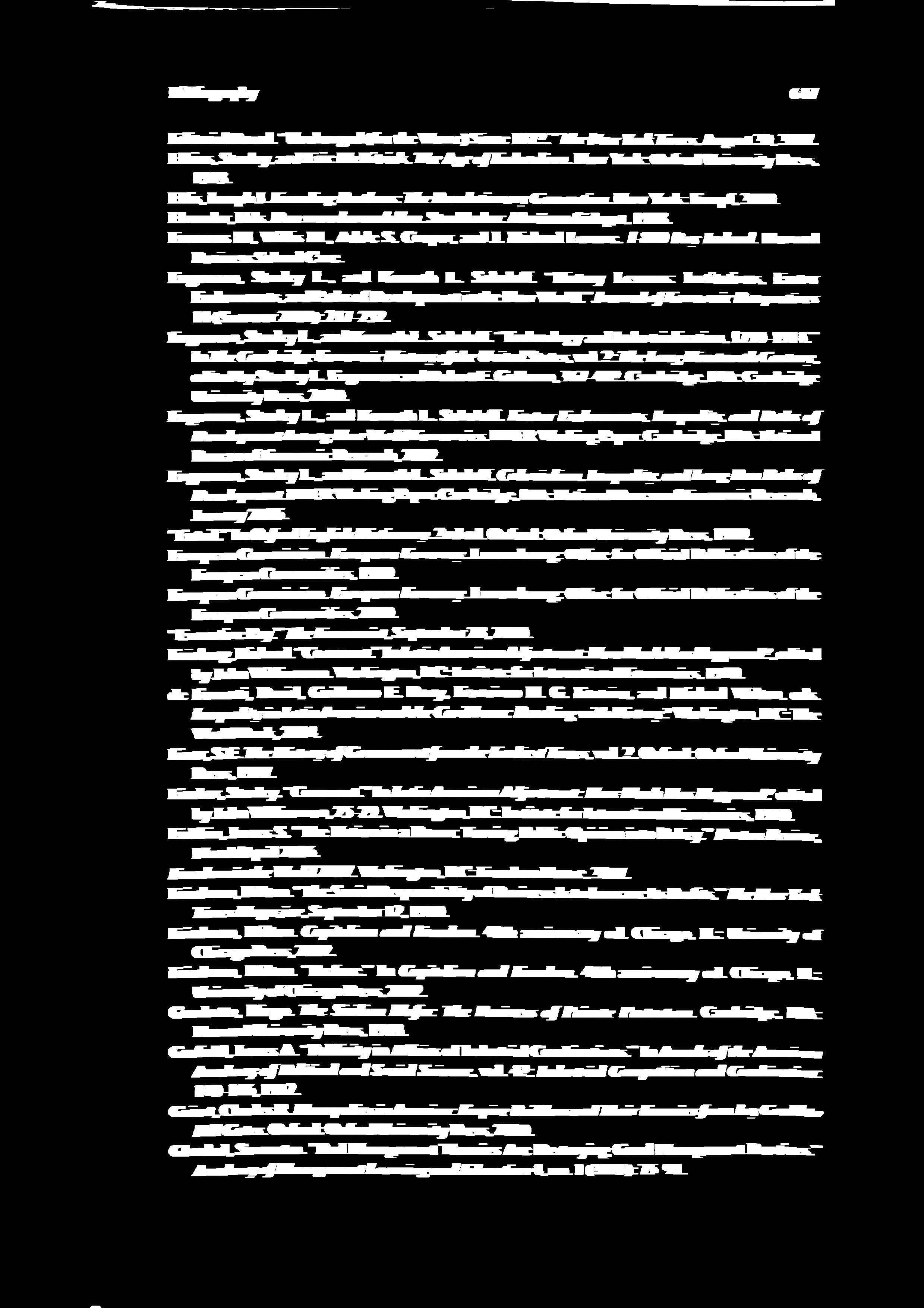}}\label{fig:dilated-example}}\hfil
\subfloat[Distance Transform]{\frame{\includegraphics[height=7cm,width=0.33\linewidth]{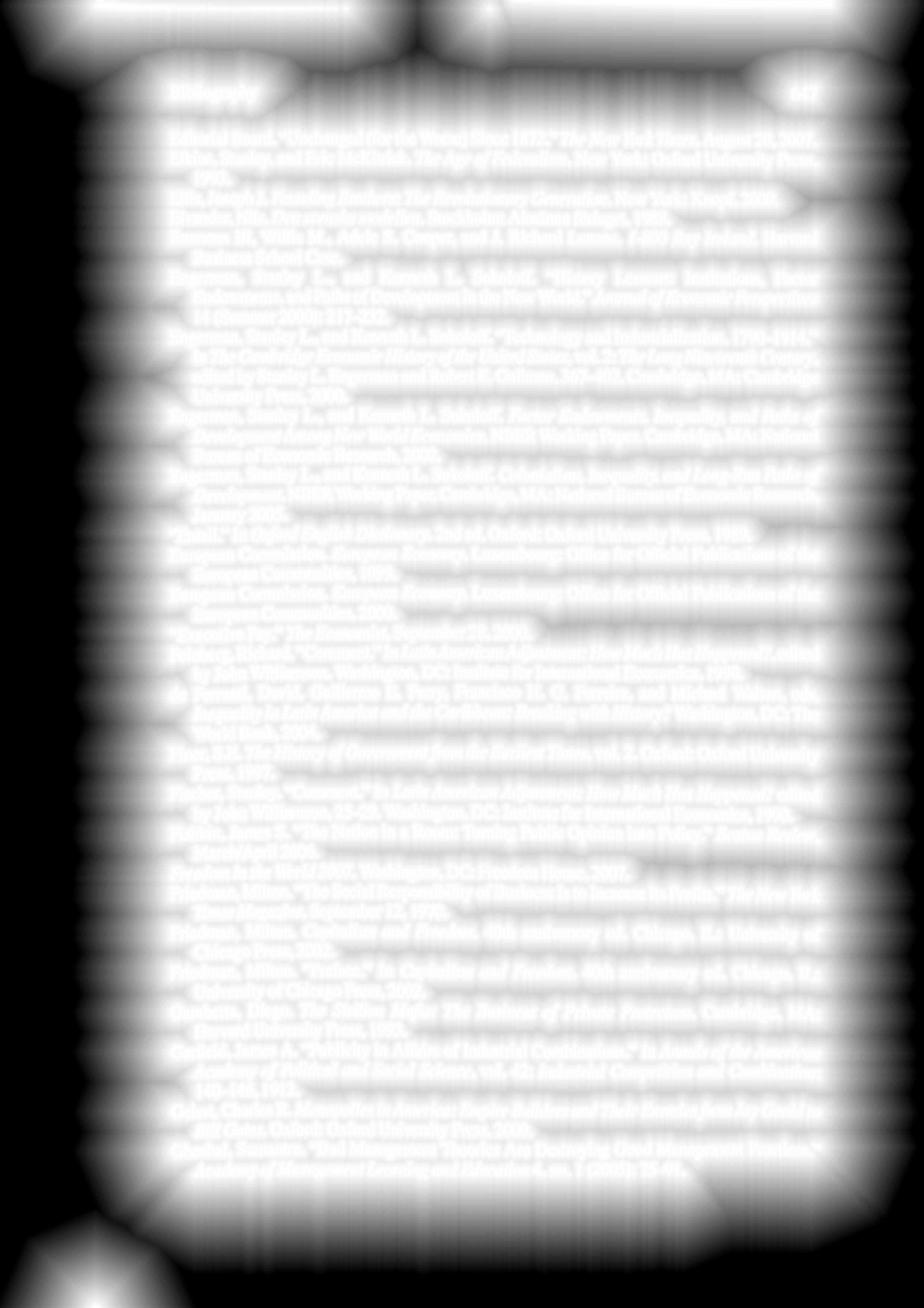}}\label{fig:dist-transform-example}}\hfil
\subfloat[Hybrid Representation]{\frame{\includegraphics[height=7cm,width=0.32\linewidth]{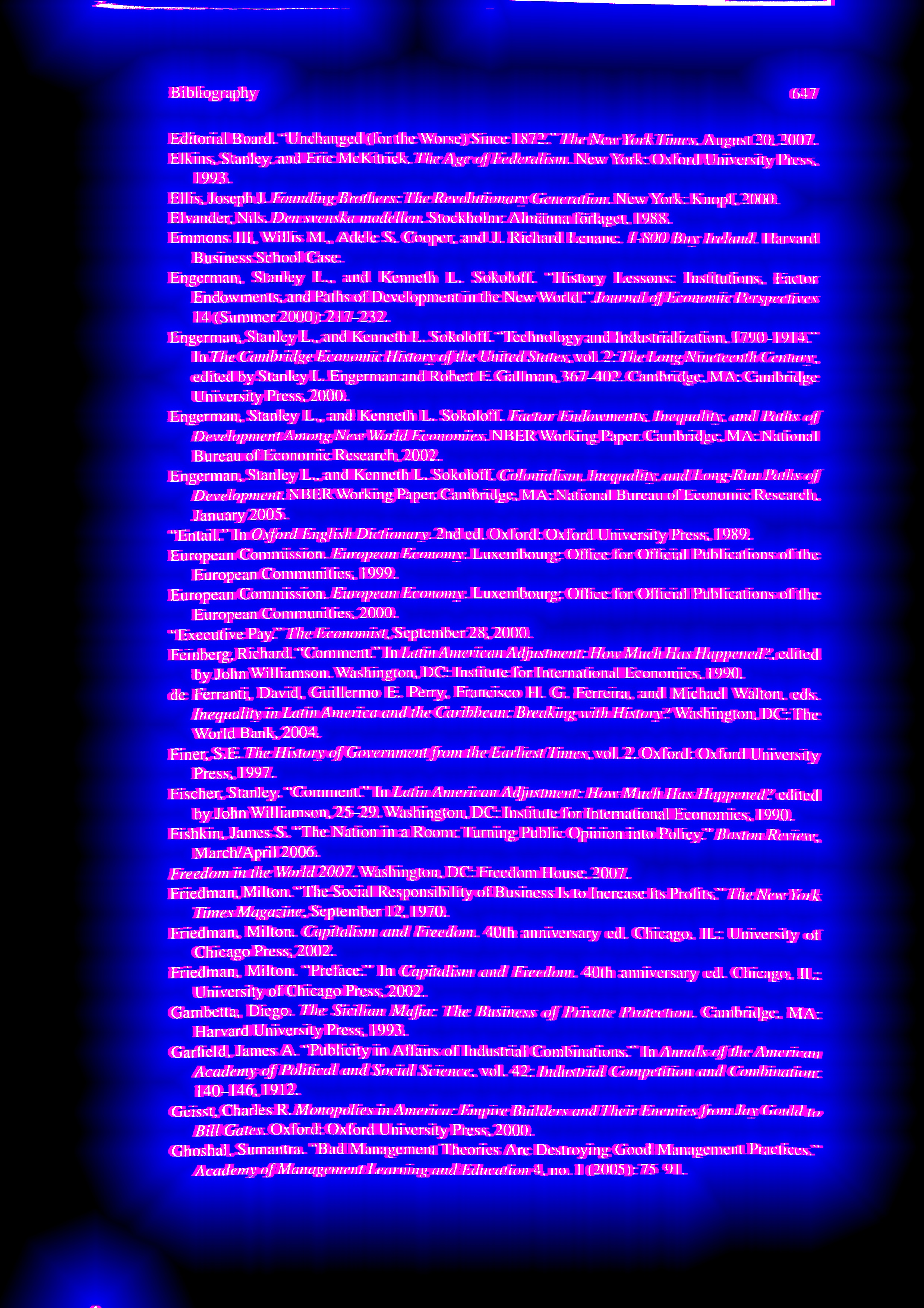}}\label{fig:preprocessed-example}}\hfil
\caption{Visual examples of a document in different pre-processing stages}
\label{fig:preprocessing_samples}
\end{figure*}

It is the final stage of the pre-processing phase, in which we merge the representations obtained from dilation and distance transform with the input image. For that purpose, we place distance transform image, binarized image, and dilated image in channels one, two, and three of an image respectively. The resultant image retains information of the original image along with additional highlighted text lines and proximity information encoded into one image. In the hybrid representation, blue color represents the proximity of the text and separation between words and line. While red color represents the color of the line. This image is later used to identify bibliographic references from a given document image. An example of the final image is shown in Fig \ref{fig:preprocessed-example}.

\subsection{Reference Detection Model}
\label{sec:reference-detection}
This section provides insights into the architecture design of the \textit{DeepBiRD}.

\subsubsection{Architecture}
\label{sec:layout-architecture2}
For the reference detection task, due to the proximity of references a network was needed which can also separate references from each other in addition to detecting those references. For that purpose, we employed a deep neural network-based architecture known as Mask {R-CNN} \cite{8237584}. It is one of the most popular networks for object detection and instance segmentation. 
The task of reference detection using layout features is a challenging task, as the references are a few pixels apart. So the detection task requires high precision to the level of each and every pixel. In contrast to the Faster-{RCNN}\cite{NIPS2015_5638}, the Mask {R-CNN} \cite{8237584}
is equipped with the ROIAlign which is a non-quantized  operation and therefore preserves the data. This resulted in more accurate detections to the pixel level. It served as one of the main reason to employ  Mask {R-CNN}\cite{8237584}
for the task in hand.

\begin{table}
\centering
\caption{Architecture details of the feature backbone ResNet-50~\cite{7780459}}
\label{tab:architecture}
\[\arraycolsep=8.4pt\def\arraystretch{1.2}
\begin{array}{ccc}
\noalign{\smallskip}\hline\noalign{\smallskip}
\text{Layer}       & \text{Output}     &   \text{Structure}   \\
\noalign{\smallskip}\hline\noalign{\smallskip}
\text{conv1}       & 112\times112      &   7\times7\text{, 64, stride2}   \\
\noalign{\smallskip}\hline\noalign{\smallskip}
                   &                   &   3\times3\text{, maxpool, stride2}   \\
\text{conv2\_x}    & 56\times56        &   \begin{bmatrix}  1\times1\times64  \\  3\times3\times64  \\ 1\times1\times256   \end{bmatrix}\times3   \\
\noalign{\smallskip}\hline\noalign{\smallskip}
\text{conv3\_x}    & 28\times28        &   \begin{bmatrix}  1\times1\times128 \\  3\times3\times128 \\ 1\times1\times512   \end{bmatrix}\times4   \\
\noalign{\smallskip}\hline\noalign{\smallskip}
\text{conv4\_x}    & 14\times14        &   \begin{bmatrix}  1\times1\times256 \\  3\times3\times256  \\ 1\times1\times1024 \end{bmatrix}\times6   \\
\noalign{\smallskip}\hline\noalign{\smallskip}
\text{conv5\_x}    & 7\times7          &   \begin{bmatrix}  1\times1\times512 \\  3\times3\times512  \\ 1\times1\times2048  \end{bmatrix}\times3   \\
\noalign{\smallskip}\hline\noalign{\smallskip}
            & 1\times1          &   \text{averagepool, fc, softmax}   \\
\noalign{\smallskip}\hline\noalign{\smallskip}
\end{array}\]
\end{table}

In our experiments we used standard ResNet-50~\cite{7780459} backbone for feature extraction. Table~\ref{tab:architecture} shows the details of ResNet-50~\cite{7780459} . Following the original implementation in \cite{Detectron2018}, we followed the original parameters of \cite{8237584} to train the network with batch normalization enabled. We used the pre-trained model ResNet-50 \cite{7780459} to initialize the network. Then it was fine-tuned on the train set of \textit{BibX} dataset with $287$ images containing $5741$ references, using transfer learning. For fine-tuning, we froze the first two blocks conv1 and conv2\_x while leaving all the remaining blocks trainable.

\subsubsection{Parameters}
\label{sec:parameters2}
The network was trained for 50 epochs with a base learning rate of 0.001. The learning rate was decreased in steps by a factor of $0.0001$ at $12$, $25$, and $37$ epochs respectively. In all experiments, the number of images per batch was set to 1.

\subsubsection{Inference}
\label{sec:inference2}
By performing inference on the input image we get coordinates of the detected reference's box along with a confidence score. The confidence score ranges between 0 and 1, where 0 being lowest and 1 being highest. It represents the extent to which the network is sure about that specific detection. Each detection in the results represents a reference. Once detection results are ready, OCR\footnote{https://github.com/tesseract-ocr/tesseract} is performed on each detected reference, thus extracting all references from an input image.

\section{Experiments \& Results}
\label{experiments-results}
To evaluate our system, we performed various experiments using \textit{DeepBIBX} \cite{10.1007/978-3-319-70096-0_30} model and multiple settings of \textit{DeepBiRD} on two publicly available datasets. These datasets include our own \textit{BibLy} \cite{NeuezugangDB} dataset and the one proposed by \textit{DeepBIBX} \cite{10.1007/978-3-319-70096-0_30}, here referred to as \textit{BibX} dataset \cite{ICONIP}. Due to the limited number of samples in \textit{BibX}, the authors augmented the whole dataset followed by resizing every image in train, validation and test set. In this section, we will elaborate the results of the experiments performed for evaluation.

\begin{table*}
\centering
\caption{Detection results from all variations of \textit{DeepBiRD} and \textit{DeepBIBX} \cite{10.1007/978-3-319-70096-0_30} on both datasets}
\label{overall_results}
\begin{tabular}{cccccccc}
\noalign{\smallskip}\hline\noalign{\smallskip}
                & Model     & Trained on        & Tested on     & mAP[0.5:0.95] & AP50      & AP75      & AR  \\
\noalign{\smallskip}\hline\noalign{\smallskip}
Experiment 1    & DeepBiRD  & BibX              & BibX          & 76.52         & 97.59     & 88.52     & 80.40 \\
                & DeepBIBX \cite{10.1007/978-3-319-70096-0_30}  & BibX              & BibX          & 32.51         & 54.22     & 36.24     & 23.28 \\
\noalign{\smallskip}\hline\noalign{\smallskip}
Experiment 2    & DeepBiRD  & BibX              & BibLy      & 64.53         & 89.40     & 75.20     & 70.50 \\
                & DeepBIBX \cite{10.1007/978-3-319-70096-0_30}  & BibX              & BibLy      & 29.03         & 52.56     & 30.27     & 21.44 \\
\noalign{\smallskip}\hline\noalign{\smallskip}
Experiment 3    & DeepBiRD  & BibX              & BibLy      & 64.53         & 89.40     & 75.20     & 70.50 \\
                & DeepBiRD  & BibX + BibLy   & BibLy      & 83.40         & 98.56     & 95.39     & 86.60 \\
\noalign{\smallskip}\hline\noalign{\smallskip}
\end{tabular}
\end{table*}

\begin{table*}
\centering
\caption{Results from ablation study of \textit{DeepBiRD} on both datasets}
\label{table:ablation-study-results}
\begin{tabular}{llcccc}
\hline\noalign{\smallskip}
Dataset     & Pre-processing Type               & mAP[0.5:0.95] & AP50    & AP75    & AR \\
\noalign{\smallskip}\hline\noalign{\smallskip}
BibX        & Dilation + Distance Transform     & 76.52         & 97.59   & 88.52   & 80.40 \\
            & Dilation                          & 75.91         & 96.90   & 88.29   & 79.80 \\
            & No Pre-processing                 & 75.28         & 96.85   & 87.32   & 79.40 \\
\noalign{\smallskip}\hline\noalign{\smallskip}
BibLy    & Dilation + Distance Transform     & 83.40         & 98.56   & 95.39   & 86.60 \\
            & Dilation                          & 83.24         & 98.54   & 95.35   & 86.50 \\
            & No Pre-processing                 & 83.16         & 98.50   & 95.30   & 86.30 \\
\noalign{\smallskip}\hline
\end{tabular}
\end{table*}

\subsection{Evaluation}
\label{sec:quantitative}

In this section, we will discuss the evaluation results of all the experiments performed to compare \textit{DeepBiRD} model with \textit{DeepBIBX} in different settings.

\subsubsection{Experiment 1: Baseline comparison of our approach with \textit{DeepBIBX} \cite{10.1007/978-3-319-70096-0_30}}
\label{sec:exp1}
The purpose of this experiment was to validate the effectiveness of our approach on \textit{BibX} dataset and compare its performance with \textit{DeepBIBX} \cite{10.1007/978-3-319-70096-0_30}. In this experiment, we trained \textit{DeepBiRD} on \textit{BibX} dataset with aforementioned parameters. We also trained a Fully Convolutional Network (FCN) \cite{7298965} on non-augmented \textit{BibX} dataset with exactly same settings as mentioned in the \textit{DeepBIBX} original paper \cite{10.1007/978-3-319-70096-0_30}. The difference being that in our experiments we used non-augmented dataset to enable results to be directly comparable with our approach. Additionally, we resized the training, validation or test set images and blurred its lines as mentioned in \cite{10.1007/978-3-319-70096-0_30}. However, It is worth mentioning that using non-augmented dataset, will result in different evaluation results from \textit{DeepBIBX}~\cite{10.1007/978-3-319-70096-0_30}. Once the training finished, both models were evaluated on non-augmented test set of \textit{BibX} dataset. By doing so it enabled us to directly compare the performance of our approach with \textit{DeepBIBX} \cite{10.1007/978-3-319-70096-0_30} approach on \textit{BibX} dataset.

Each detection was validated using its Intersection over Union (IoU) with ground truth annotations. Both models were evaluated on different IoU thresholds ranging from 0.50 to 0.95 which is a standard for an object detection problem. A detection is considered as correct detection if the IoU of a given bounding box is greater than the IoU threshold. Table \ref{overall_results} shows comparison of \textit{DeepBiRD} results with \textit{DeepBIBX} \cite{10.1007/978-3-319-70096-0_30} for experiment 1. The results show that \textit{DeepBiRD} was able to achieve an average precision and average recall of $76.52\%$ and $80.40\%$ respectively. On the other hand \textit{DeepBIBX} was only able to achieve an average precision and average recall of $32.51\%$ and $23.28\%$ respectively. Even at the lowest IoU threshold of $0.50$, \textit{DeepBiRD} was able to perform significantly better even more than a factor of 2.

The reason behind the strong performance of \textit{DeepBiRD} is that it is based on Mask {R-CNN} \cite{8237584} which performs semantic segmentation on shortlisted ROIs on the other hand FCN performs semantic segmentation on complete image.

\subsubsection{Experiment 2: Robustness}
\label{sec:exp2}
The purpose of this experiment was to validate the extent of robustness for both \textit{DeepBiRD} and \textit{DeepBIBX} \cite{10.1007/978-3-319-70096-0_30}. To do so, we evaluated both systems on more unseen data i.e. test set from another dataset.
The \textit{DeepBiRD} and \textit{DeepBIBX} models trained in the Experiment 1 were reused in this experiment.
Both models were trained on Non-Augmented \textit{BibX} dataset and evaluated on test set of \textit{BibLy} dataset. The results from this experiment show the extent of effectiveness of \textit{DeepBIBX} \cite{10.1007/978-3-319-70096-0_30} \& \textit{DeepBiRD} on unseen data. Both models were evaluated on a range of IoU thresholds ranging from $0.50$ to $0.95$. Table \ref{overall_results} shows the evaluation results of \textit{DeepBIBX} \cite{10.1007/978-3-319-70096-0_30} model on \textit{BibLy} dataset. 
The results show that the performance of both models slightly decreased as expected when they are applied to unseen data. \textit{DeepBiRD} was able to achieve an average precision and average recall of $64.53\%$ and $70.50\%$ respectively. On the other, \textit{DeepBIBX} was able to achieve an average precision and average recall of $29.03\%$ and $21.44\%$ respectively. Therefore, outperforming \textit{DeepBIBX} by a significant margin similar to experiment 1 results.

\subsubsection{Experiment 3: Generalization}
\label{sec:exp3}
The purpose of this experiment was to verify \textit{DeepBiRD} for generalization by employing transfer learning to adapt the network to the \textit{BibLy} dataset. In this experiment, the pre-trained \textit{DeepBiRD} model on \textit{BibX} dataset was used as a baseline and was then fine-tuned on the train set of \textit{BibLy} dataset to learn more reference examples. Once the training was finished, the final model was evaluated against the baseline model on the test set of \textit{BibLy} dataset.

The results of this experiment are shown in Table \ref{overall_results}. The fine-tuned model was able to achieve an average precision and average recall of $83.40\%$ and $86.60\%$ respectively, which is $18.87\%$ and $16.1\%$ better than the results before fine-tuning of the model. However, precision at IoU of $0.50$ and $0.75$ increased by $9.16\%$ and $20.19\%$ respectively. This indicates that after fine-tuning, the model was significantly improved and was able to detect bibliographic references with a higher overlap. From these results, we can infer that \textit{DeepBiRD} can be generalized as it can adapt very well to new data.

\subsection{Ablation Study for Input Representation}
\label{sec:ablation-study}
This section discusses the results of the ablation study to show the impact of hybrid representation used in  \textit{DeepBiRD}. The purpose of this analysis was to determine the effectiveness of individual components in the pre-processing phase. For this analysis, we designed several experiments with different pre-processing configurations. In the first experiment, we employed the aforementioned pre-processing steps i.e. distance transform, dilation, and merging them with the original input image. In the second experiment, we employed dilation as a sole step in the pre-processing phase. Lastly in the third experiment, we excluded the pre-processing phase and used the original input image without any pre-processing. These experiments will highlight the contribution of individual components in the pre-processing phase towards the final output.

We used \textit{BibX} dataset to perform this ablation study. Table \ref{table:ablation-study-results} shows the results of different representation types employed in various experiments. The evaluation results show that the experiment which employed both dilation and distance transform along with merging channels sets the baseline average precision and an average recall of $76.52\%$ and $80.40\%$ respectively. In the second experiment, pre-processing consisted of dilation of the input image. This resulted in a decrease of $0.6\%$ in both average precision and average recall, therefore suggesting that providing distance transform aided the proposed system to detect bibliographic references from a given document image. In the third experiment, the pre-processing phase was removed altogether and the original input image was fed to the system with no pre-processing. This resulted in a further decrease in average precision and average recall by $0.63\%$ and $0.40\%$ respectively. Therefore suggesting that dilation also contributed towards improving system performance.

To verify the trend in results, we performed the same ablation study on a second dataset \textit{BibLy} dataset. Table \ref{table:ablation-study-results} shows the results of this analysis. In the first experiment, with dilation and distance transform as a part of pre-processing, it sets the baseline evaluation average precision and average recall of $83.40\%$ and $86.60\%$ respectively. In the second experiment with pre-processing involving dilation of the input image decreased the average precision and average recall by $0.16\%$ and $0.10\%$ respectively. Whereas in the third experiment with no pre-processing, the average precision and average recall were further decreased by $0.08\%$ and $0.20\%$ respectively. These trends in results proved to be consistent that both dilation and distance transform play their part in further improving the performance of the system.

\begin{table}
\centering
\caption{Performance comparison of \textit{DeepBiRD} with other object detection models on BibX dataset}
\label{table:sota-results}
\begin{tabular}{lcc}
\hline\noalign{\smallskip}
Model                                                                            & AP50(\%)     & Diff.(\%) \\
\noalign{\smallskip}\hline\noalign{\smallskip}
DeepBiRD (Proposed Method)                                                       & 96.85        & -      \\
Faster R-CNN \cite{NIPS2015_5638}\cite{dicta19Ref}                               & 84.50        & -12.35 \\
Deformable Faster R-CNN \cite{dai17dcn}\cite{NIPS2015_5638}\cite{dicta19Ref}     & 82.83        & -14.02 \\
Deformable RFCN \cite{dai17dcn}\cite{dai16rfcn}\cite{dicta19Ref}                 & 82.37        & -14.48 \\
Deformable FPN \cite{dai17dcn}\cite{lin2017feature}\cite{dicta19Ref}             & 84.17        & -12.68 \\
\noalign{\smallskip}\hline
\end{tabular}
\end{table}

\begin{figure*}[!t]
\centering
  \includegraphics[width=0.80\textwidth]{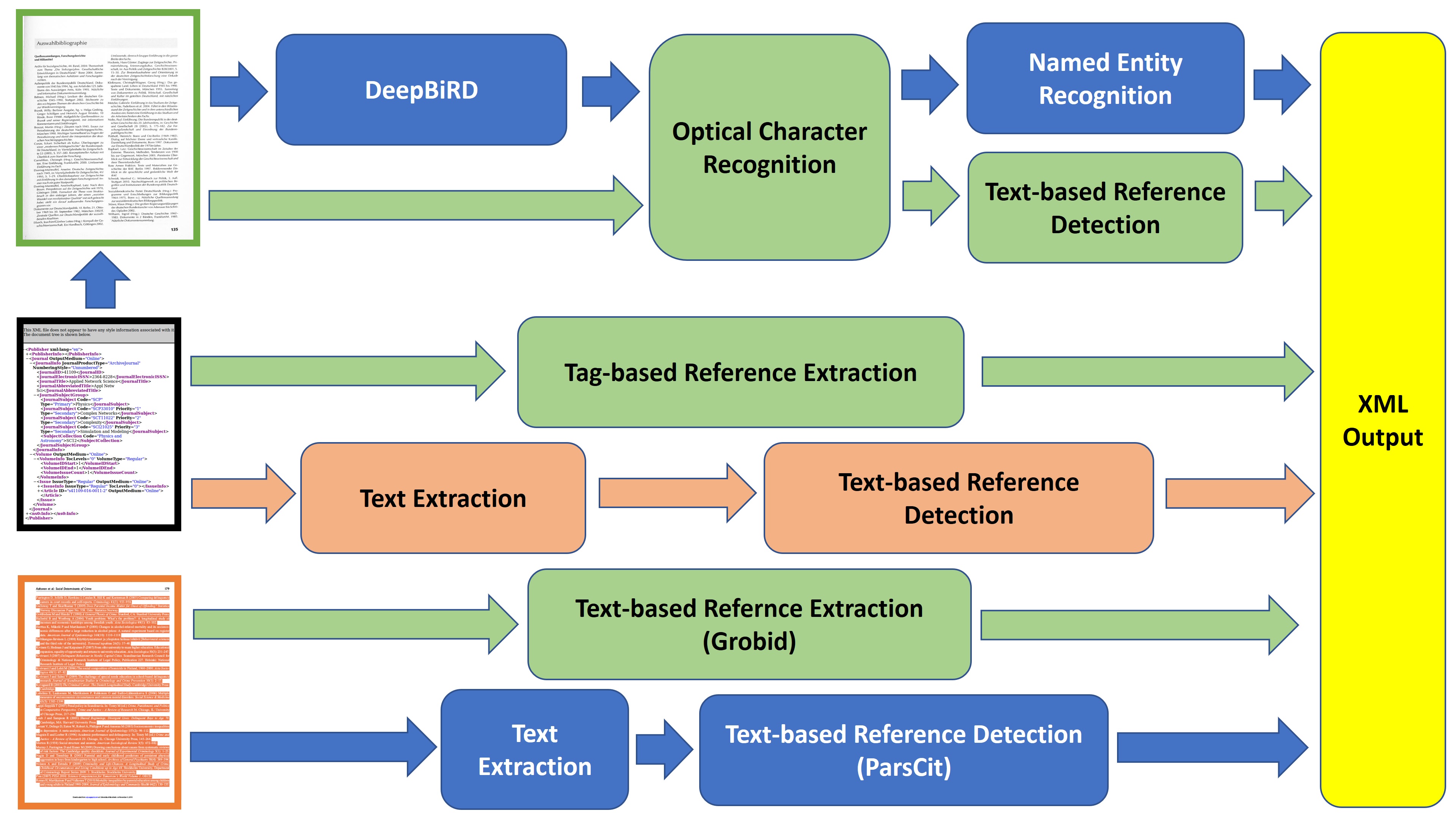}
\caption{Overview of the BRExSys}
\label{fig:overallPipeline}
\end{figure*}

\subsection{Overall discussion}
\label{sec:overall-discussion}
In this paper, we presented a new layout driven reference detection approach called "\textit{DeepBiRD}" which exploits human intuition and visual cues to effectively detect references without taking textual features into account. By meticulous experimentation, we pushed the boundaries of automatic reference detection and set a new state-of-the-art. The evaluation results clearly show that the proposed approach \textit{DeepBiRD} is an effective, robust and generalized approach by outperforming \textit{DeepBIBX} \cite{10.1007/978-3-319-70096-0_30} with significant margins. For the sake of completion, we compared the performance of \textit{DeepBiRD} with some other object detection models presented in the existing literature. \cite{dicta19Ref} performed a benchmark of four object detection models on the \textit{BibX} dataset. Table~\ref{table:sota-results} shows the performance comparison between \cite{dicta19Ref} and \textit{DeepBiRD}. It is worth mentioning that we are using the results of \textit{DeepBiRD} without pre-processing to fairly compare the results as there was no pre-processing used in \cite{dicta19Ref}. Results show that our proposed approach \textit{DeepBiRD} performed on average 13.38\% better than each of the presented object detection model on \textit{BibX} dataset. Faster R-CNN \cite{NIPS2015_5638} was the best performing model among the benchmark selection. However it was 12.35\% worse than our proposed approach \textit{DeepBiRD}. One of the possible reasons of \textit{DeepBiRD} superiority is its ROIAlign operation which not only preserves data but also results in precise detection up to pixel level.

\begin{figure*}[!t]
    \centering
  \subfloat[DeepBiRD\label{dbr-best}]{%
       \frame{\includegraphics[height=5.7cm,width=0.26\linewidth]{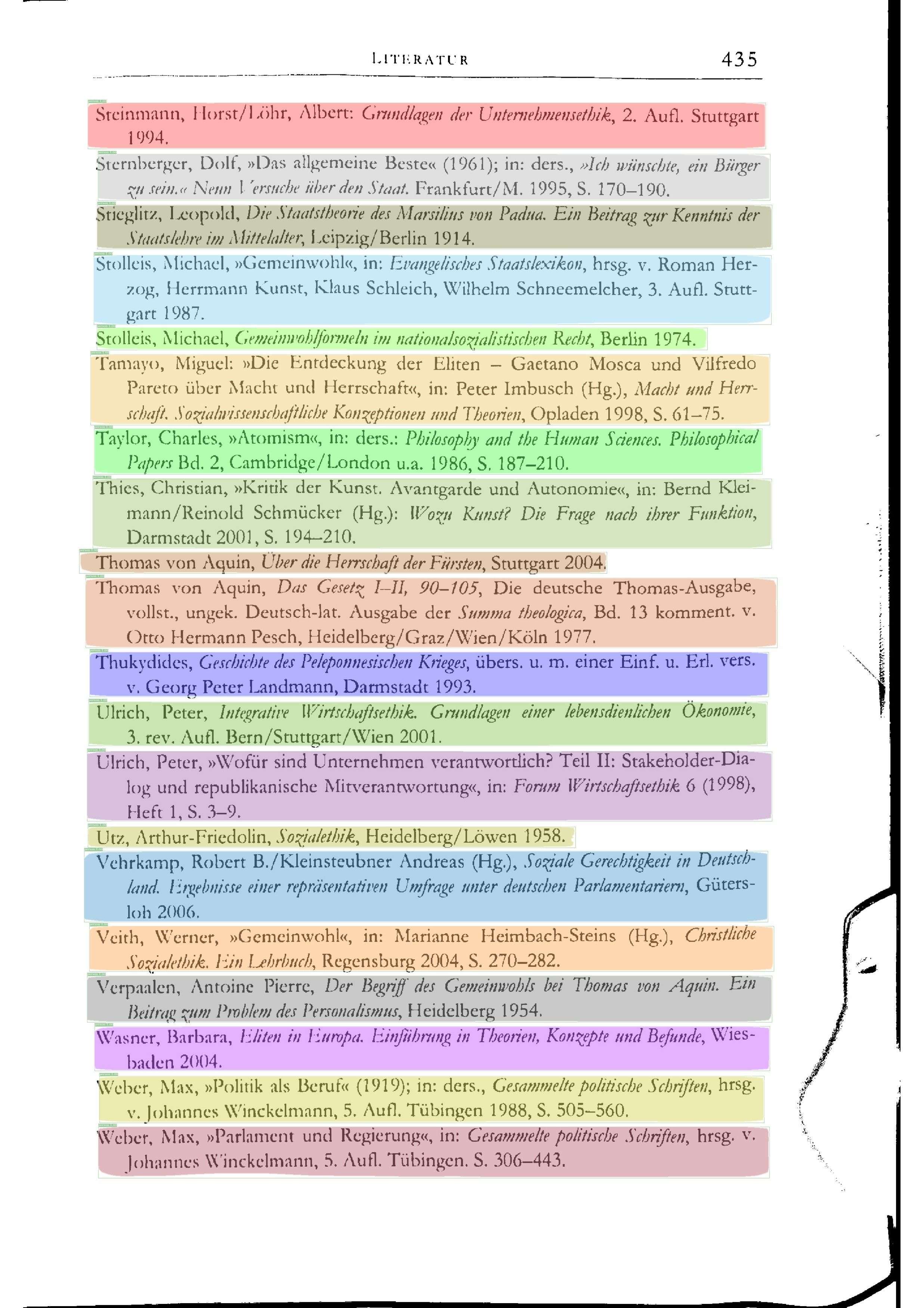}}}\hfill
  \subfloat[DeepBIBX\cite{10.1007/978-3-319-70096-0_30}\label{dbb-best}]{%
        \frame{\includegraphics[height=5.7cm,width=0.26\linewidth]{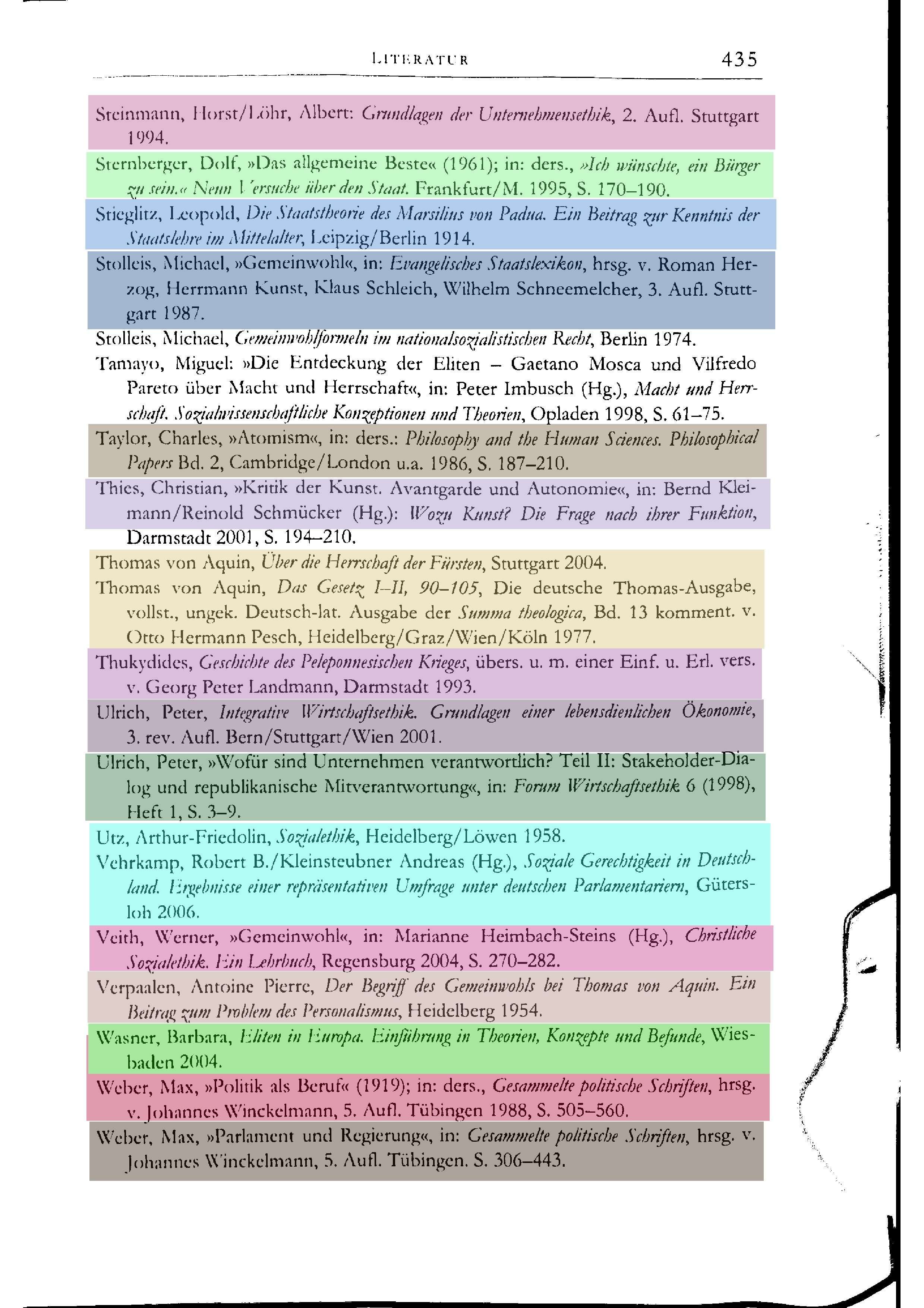}}}\hfill
  \subfloat[ParsCit\cite{CouncillGK08}\label{par-best}]{%
        \frame{\includegraphics[height=5.7cm,width=0.26\linewidth]{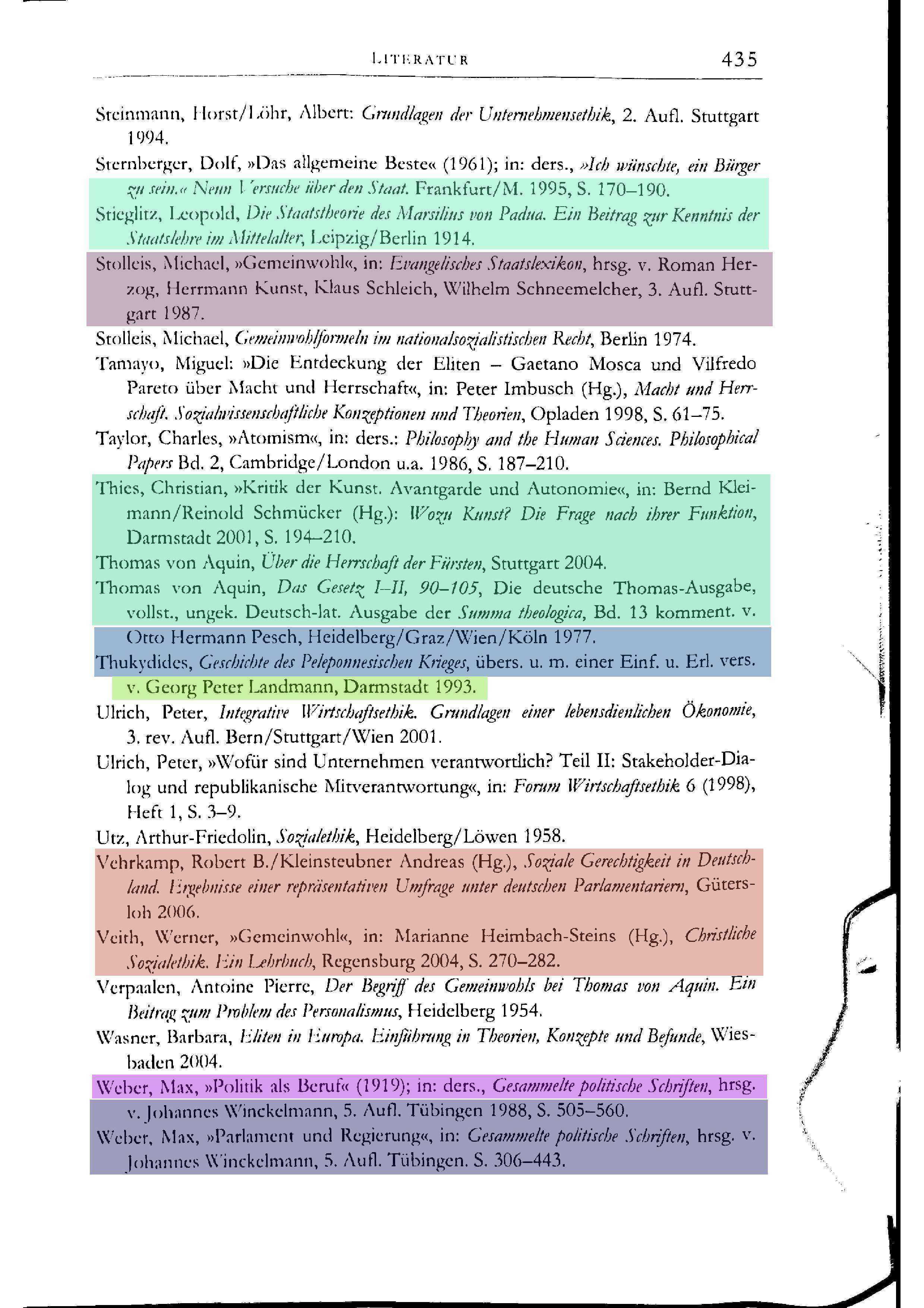}}}\hfill
        
  \caption{Best case output sample in comparison with state-of-the-art approaches}
  \label{fig:best_cases} 
\end{figure*}

\begin{figure*}[!t]
    \centering
  \subfloat[DeepBiRD\label{dbr-average}]{%
       \frame{\includegraphics[height=5.7cm,width=0.26\linewidth]{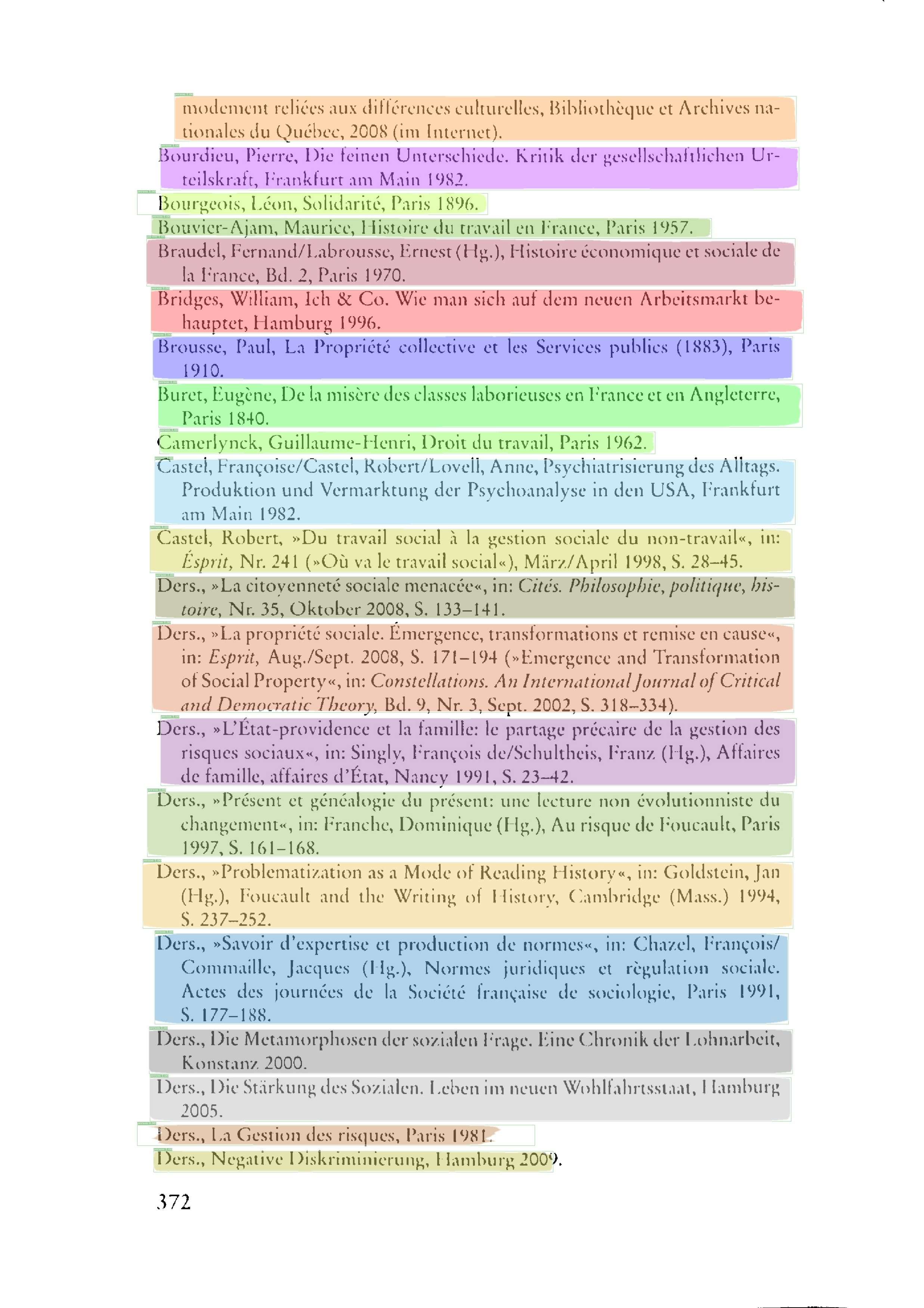}}}\hfill
  \subfloat[DeepBIBX\cite{10.1007/978-3-319-70096-0_30}\label{dbb-average}]{%
        \frame{\includegraphics[height=5.7cm,width=0.26\linewidth]{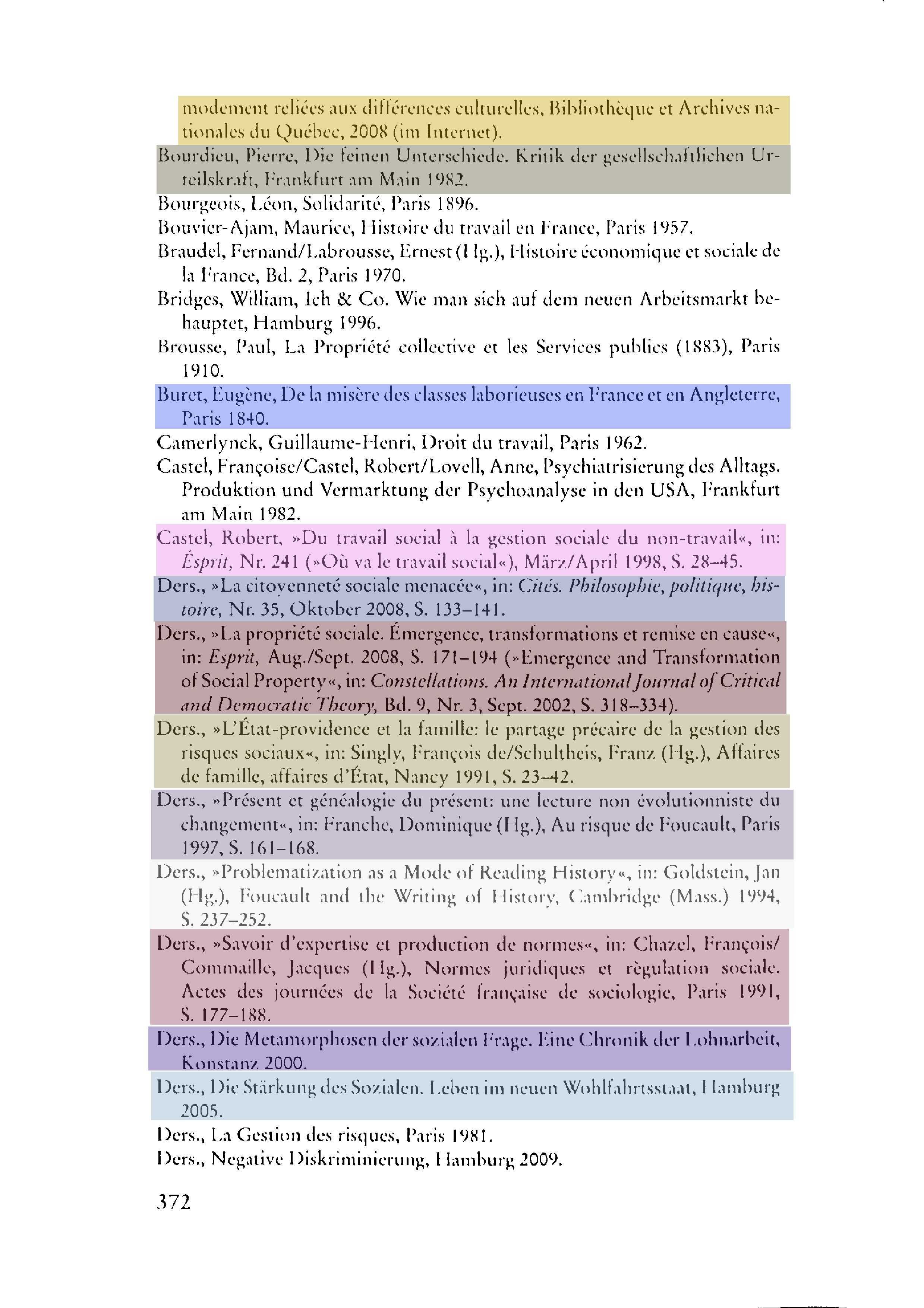}}}\hfill
  \subfloat[ParsCit\cite{CouncillGK08}\label{par-average}]{%
        \frame{\includegraphics[height=5.7cm,width=0.26\linewidth]{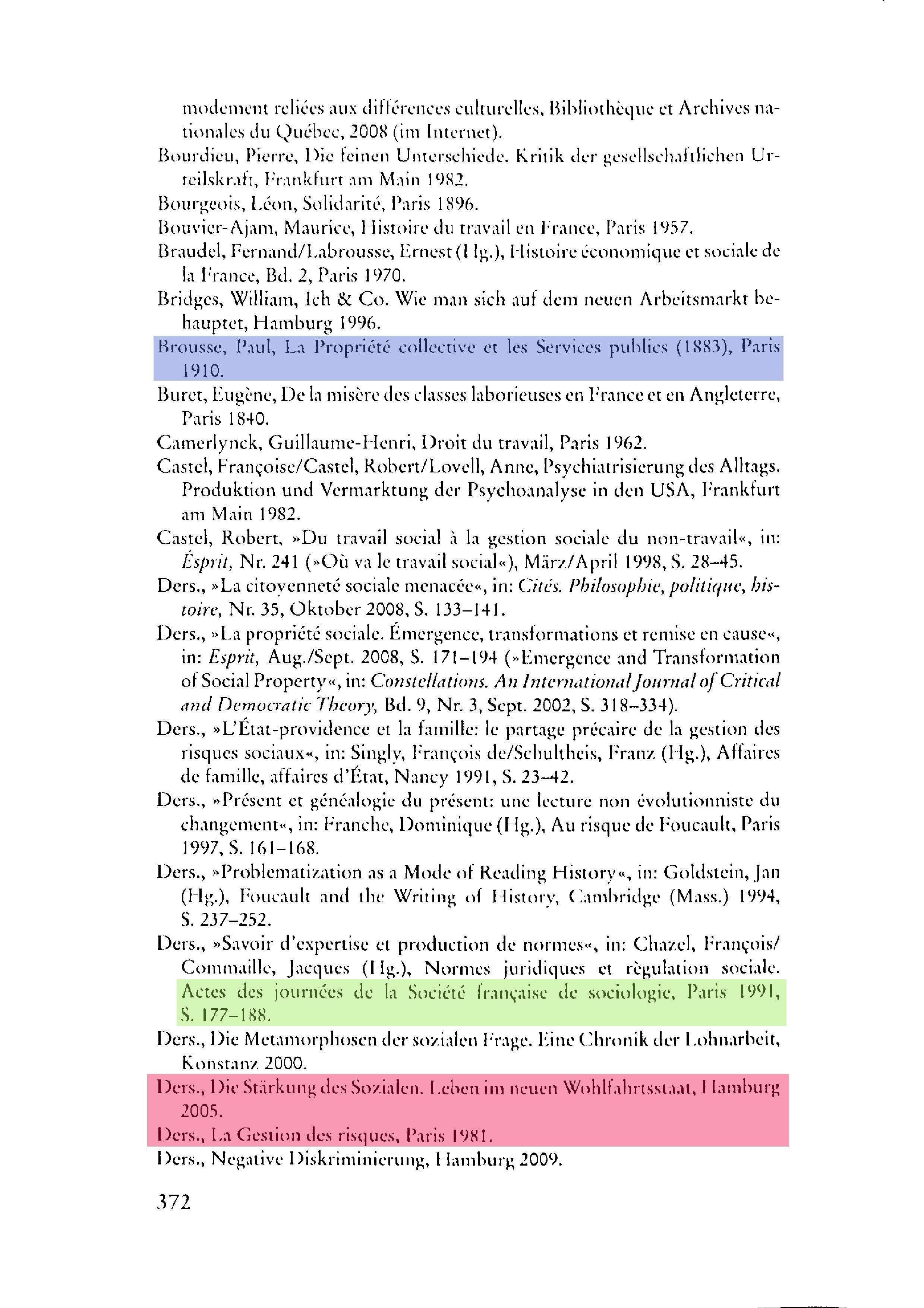}}}\hfill
        
  \caption{Average case output sample in comparison with state-of-the-art approaches}
  \label{fig:average_cases} 
\end{figure*}

\begin{figure*}[!t]
    \centering
  \subfloat[DeepBiRD\label{dbr-worst}]{%
       \frame{\includegraphics[height=5.7cm,width=0.26\linewidth]{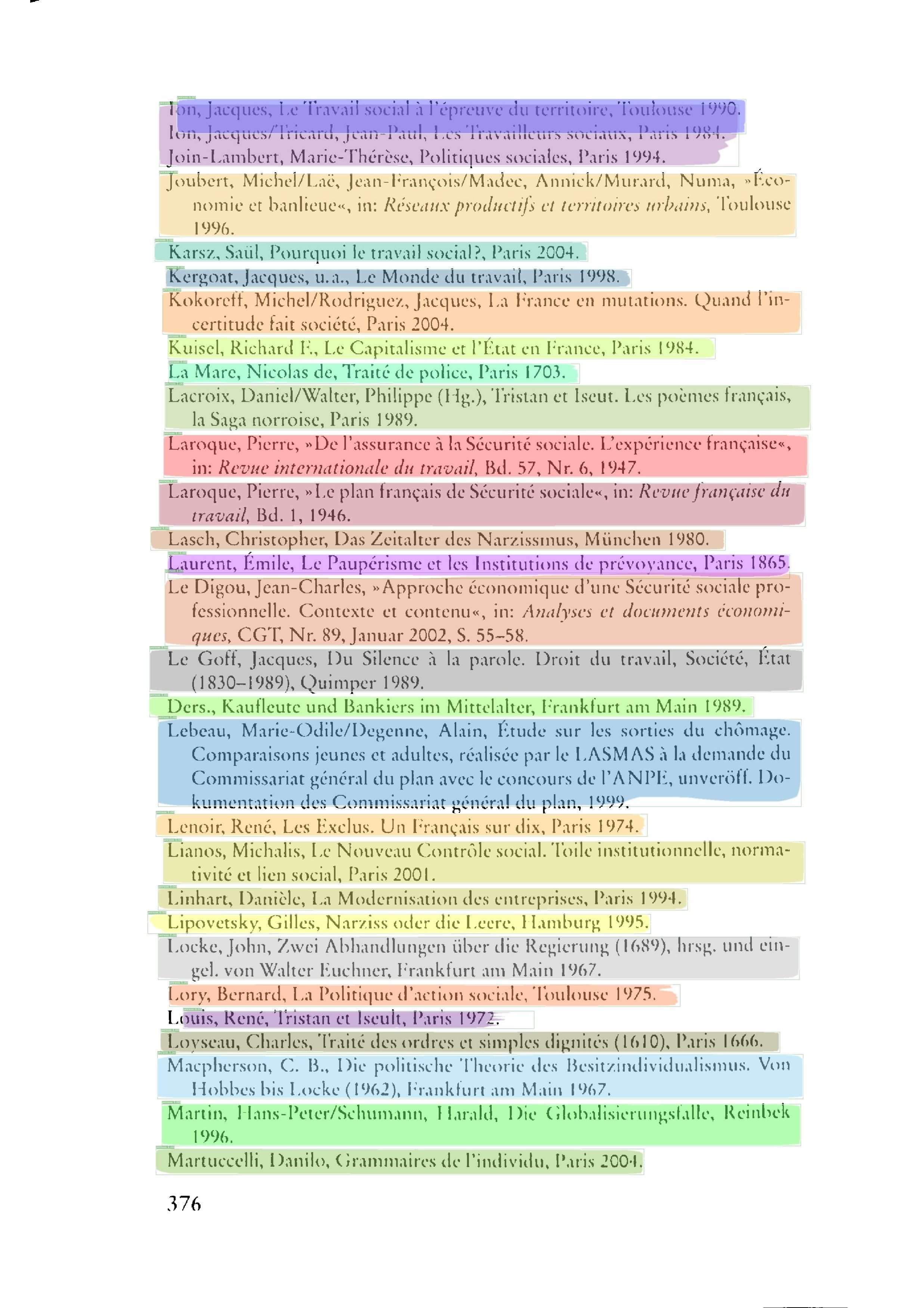}}}\hfill
  \subfloat[DeepBIBX\cite{10.1007/978-3-319-70096-0_30}\label{dbb-worst}]{%
        \frame{\includegraphics[height=5.7cm,width=0.26\linewidth]{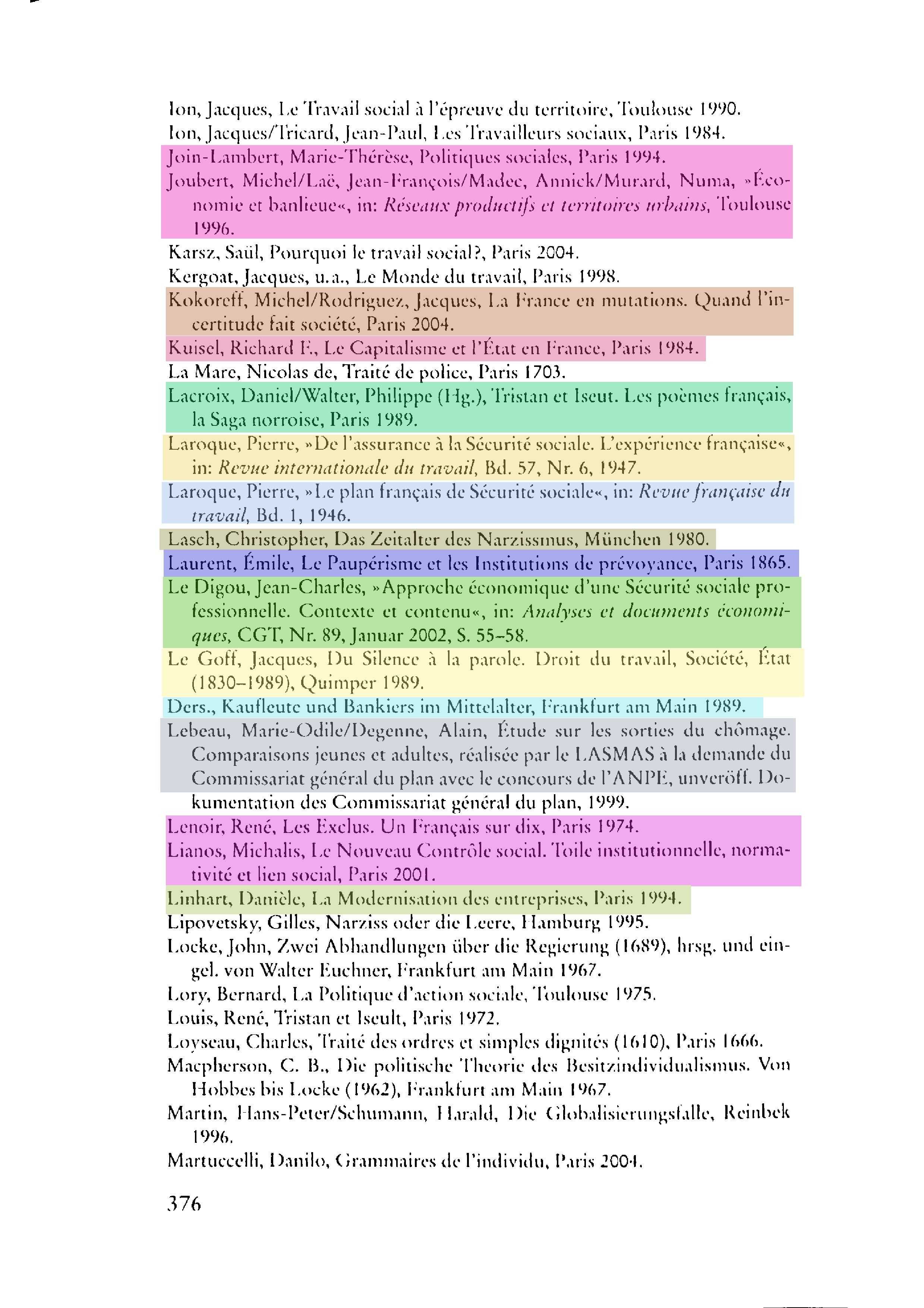}}}\hfill
  \subfloat[ParsCit\cite{CouncillGK08}\label{par-worst}]{%
        \frame{\includegraphics[height=5.7cm,width=0.26\linewidth]{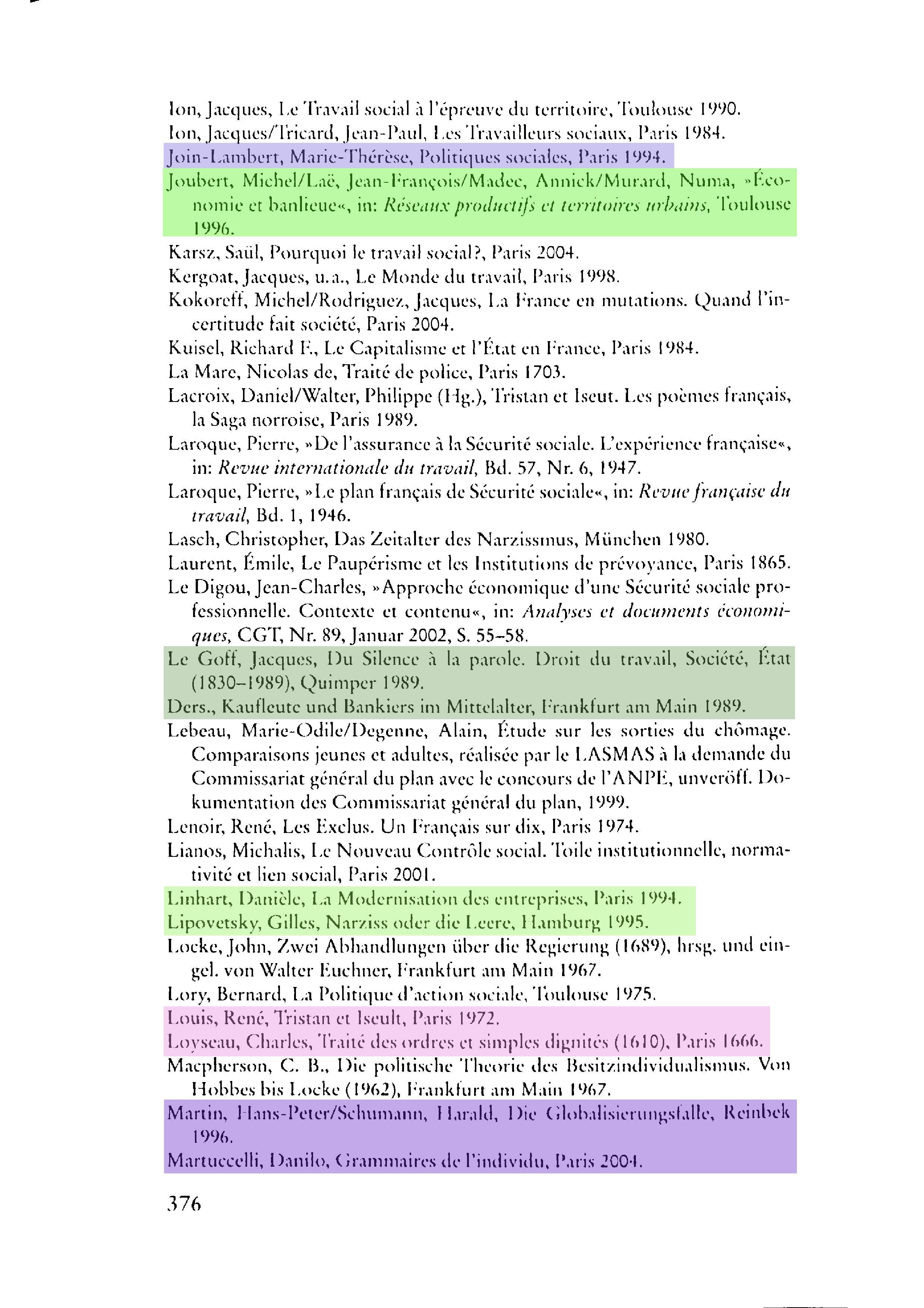}}}\hfill
        
  \caption{Worst case output sample in comparison with state-of-the-art approaches}
  \label{fig:worst_cases} 
\end{figure*}

Fig \ref{fig:best_cases}, \ref{fig:average_cases} \& \ref{fig:worst_cases} show visual examples of best, average and worst results from our system. Results from \textit{DeepBIBX} \cite{10.1007/978-3-319-70096-0_30} and text-based model \textit{ParsCit} for each example are also shown for comparison. All these results demonstrate the dominance of \textit{DeepBiRD} over all other text-based or layout-based approaches. Trained model of the above mentioned experiments is available at the URL\footnote{https://github.com/rtahseen/DeepBiRD}.


\section{BRExSys: A Bibliographic Reference Extraction System}
This section discusses the details of our proposed framework. Our framework unifies all state-of-the-art bibliographic reference detection methods in one place to detect and extract references from scanned, markup, and textual documents. To take advantage of multiple models to the full extent, we provide various possibilities to use these models individually or in a fusion. The overview of our complete system is shown in Fig \ref{fig:overallPipeline} and details of the proposed system are discussed as follows:

\subsection{Reference Extraction from Scanned Documents}
\begin{figure}
    \centering
    \begin{minipage}{0.49\textwidth}
        \centering
        \includegraphics[width=0.9\textwidth]{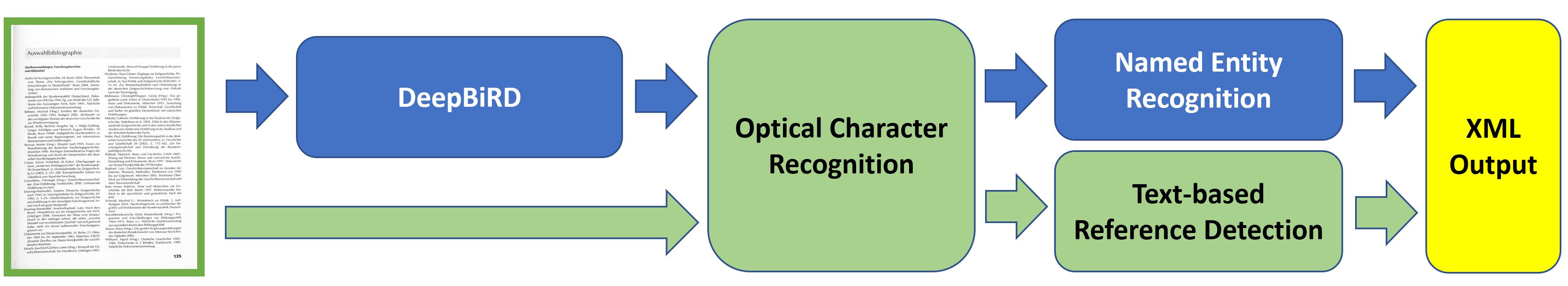} 
        \caption{Pipeline for Scanned Documents}
        \label{fig:scannedPipeline}
    \end{minipage}\hfill
    \begin{minipage}{0.49\textwidth}
        \centering
        \includegraphics[width=0.9\textwidth]{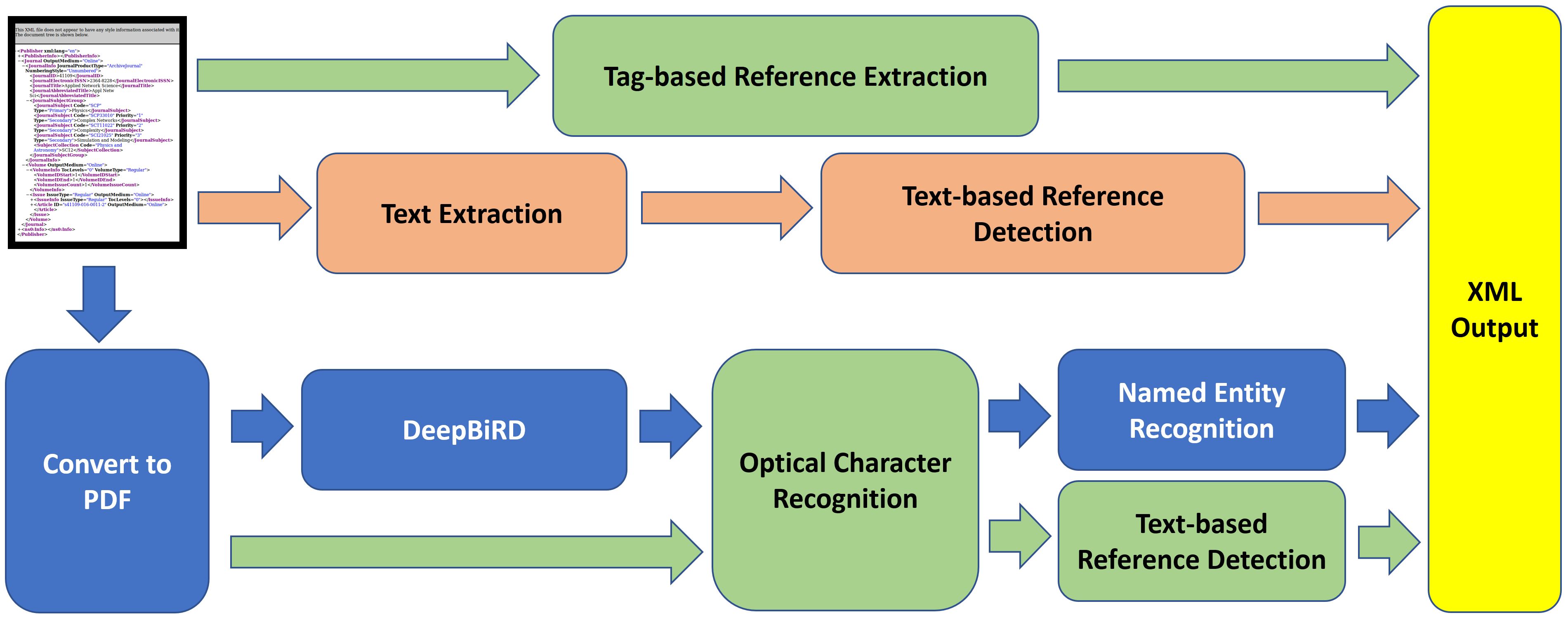} 
        \caption{Pipeline for Markup Documents}
        \label{fig:markupPipeline}
    \end{minipage}
\end{figure}

In this section, we will discuss pipelines specific for bibliographic reference extraction from scanned documents. The overview of these pipelines is shown in Fig \ref{fig:scannedPipeline}. For scanned documents, we provide two pipelines i.e. Layout-based pipeline and Text-based pipeline. The layout-based pipeline is represented in blue color while the text-based pipeline is represented in green color. A scanned document can also be processed through both pipelines simultaneously and for such cases, results from both pipelines are included in the final output XML.

\subsubsection{Layout-based pipeline}
In a layout-based pipeline, we employed \textit{DeepBiRD}, a state-of-the-art layout-driven reference extraction model. Provided a scanned document \textit{DeepBiRD} performs bibliographic reference detection on the individual document image. Lastly, we employed \textit{ParsCit} \cite{CouncillGK08} to carry out Named Entity Recognition (NER) on each detected reference to identify reference string metadata like author names, publication title, publication year, etc. All the results are eventually returned in the form of a predefined standard XML file format.

\subsubsection{Text-based pipeline}
In text-based pipeline, we extract all the text from given scanned document and use it for text-based reference extraction. For this purpose we employed \textit{ParsCit} \cite{CouncillGK08}, a state-of-the-art text-based reference extraction model. Additionally, \textit{ParsCit} \cite{CouncillGK08} extracts reference string metadata by performing NER on extracted bibliographic reference strings.

\subsection{Reference Extraction from Textual Documents}
This section discusses reference extraction pipelines from text documents like born-digital PDFs and plain text files. The pipeline overview for bibliographic reference extraction from textual documents is shown in Fig \ref{fig:textPipeline}. We provide three pipelines for extracting references and their respective metadata from a given textual document. Each of these pipelines is discussed as follows:

\subsubsection{Text-based pipeline (Grobid)}
In this pipeline, we employed \textit{Grobid} \cite{10.1007/978-3-642-04346-8_62}. It takes born-digital PDF as an input and extracts bibliographic references along with their metadata from a given PDF document. Extracted data is returned in the form of predefined XML. \textit{Grobid} does not depend upon other tools for text extraction from a given PDF document, therefore avoiding the introduction of a potential text extraction error.

\subsubsection{Text-based pipeline (ParsCit)}
In this pipeline, we firstly extract text from a given textual document and is further processed for bibliographic reference and metadata extraction. For this workflow we employed \textit{ParsCit} \cite{CouncillGK08}. It takes raw text as an input and extracts references along with its metadata from the given text.


\subsubsection{Layout-based pipeline}
Tertiary workflow serves as an alternate solution suggesting that a born-digital PDF can also be processed as a scanned document using layout driven reference extraction. In this workflow, we employed a state-of-the-art layout based reference detection and extraction approach called  \textit{DeepBiRD}. Details of this pipeline are already discussed in the section discussing reference extraction from Scanned documents.

\subsection{Reference extraction from markup documents}
In this section, we will discuss bibliographic reference extraction from markup documents like HTML and XML. Markup documents usually consist of a peculiar hierarchy. Depending on a known topology of a given markup document we provide multiple workflows for extracting bibliographic references and their metadata from markup documents. The overview of the pipelines is shown in Fig \ref{fig:markupPipeline}, where each pipeline handling a specific case is represented in a different color.

\subsubsection{Direct mapping pipeline}
The direct mapping pipeline deals with the case when we are fully aware of tags hierarchy in the markup document i.e. XML or HTML document from Zotero \cite{zotero}. In such cases we perform tag-based reference extraction by targeting relevant tags like author name, title, publisher, etc, thus extracting all references from markup document along with their metadata.

\subsubsection{Text-based pipeline}
Text-based pipeline deals with the case where we have partial knowledge about the tags hierarchy of a given markup document i.e. XML or HTML document generated from older versions of Zotero \cite{zotero}. In such cases, we first extract all the text from the markup document using all known tags and then perform text-based reference extraction on extracted text. We employ \textit{ParsCit} \cite{CouncillGK08} for extracting references and their respective metadata from the extracted text.

\begin{figure}
    \centering
    \begin{minipage}{0.49\textwidth}
        \centering
        \includegraphics[width=0.9\textwidth]{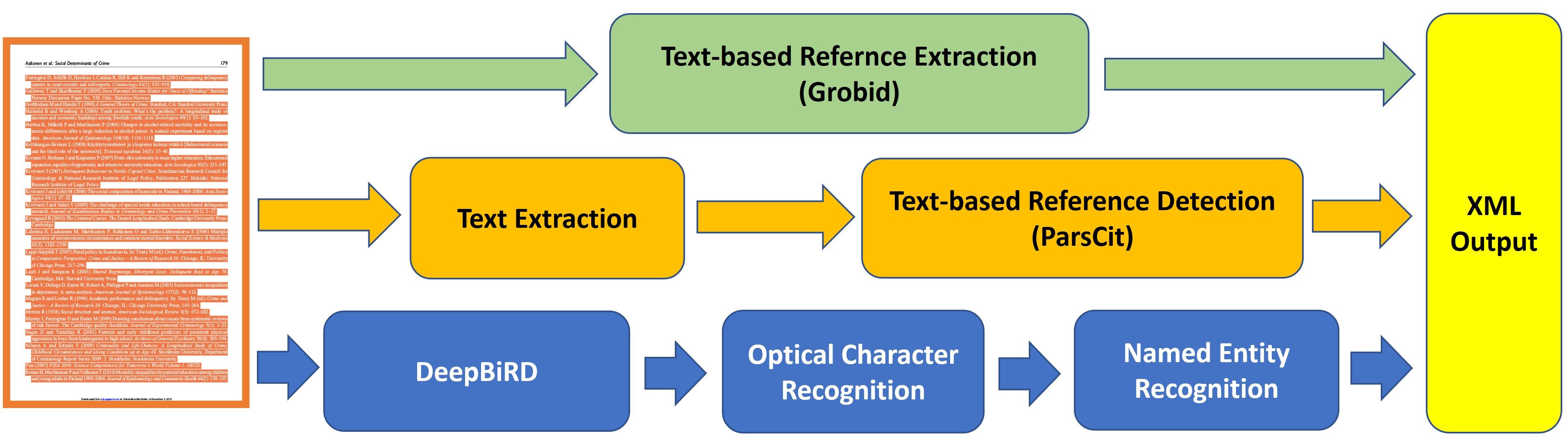} 
        \caption{Pipeline for Textual Documents}
        \label{fig:textPipeline}
    \end{minipage}\hfill
    \begin{minipage}{0.49\textwidth}
        \centering
        \includegraphics[width=0.9\textwidth]{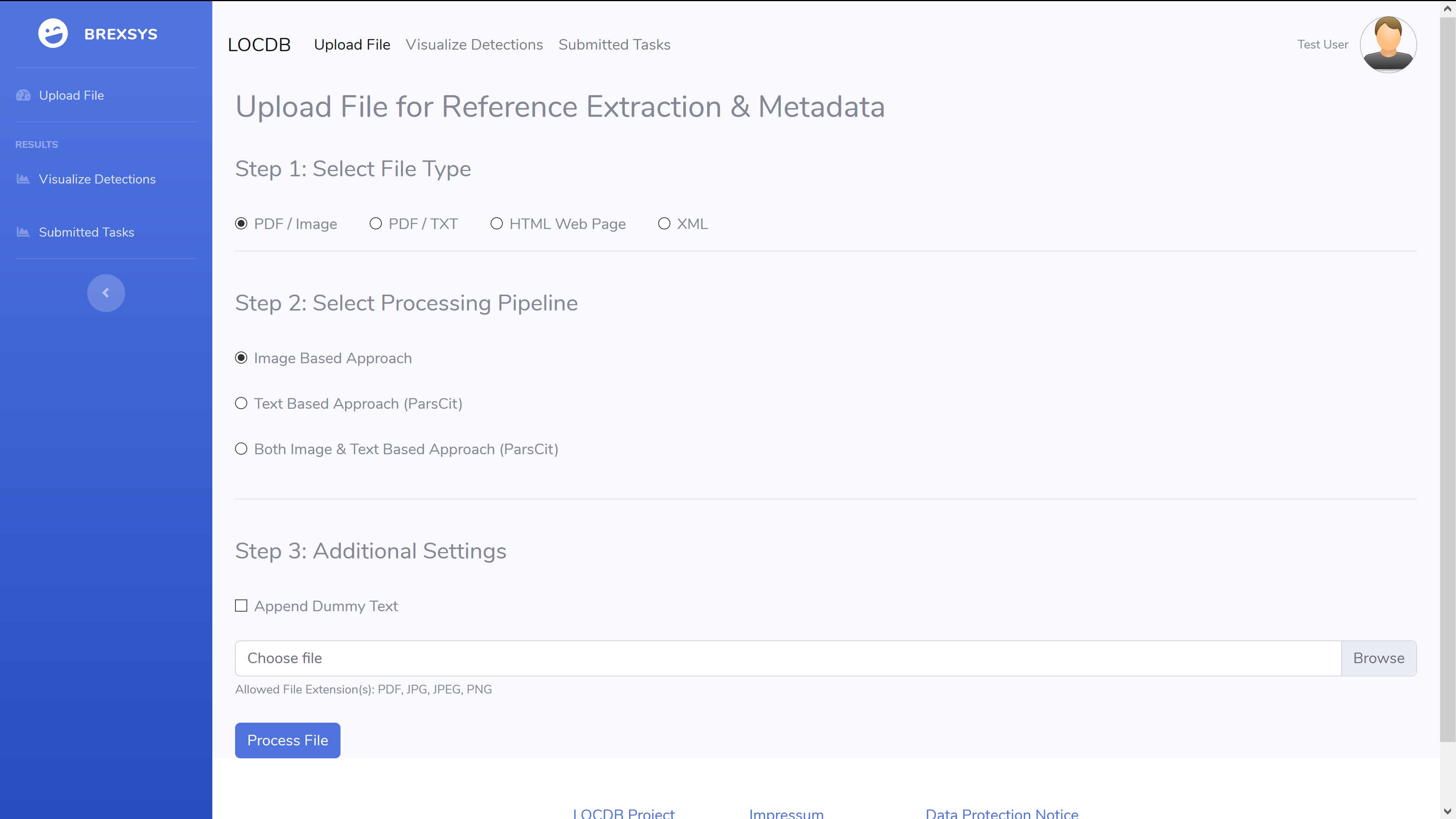} 
        \caption{Input interface}
        \label{fig:inputInterface}
    \end{minipage}
\end{figure}


\subsubsection{Layout-based pipeline}
Layout-based pipeline deals with the case when a given markup document is in HTML format and has an unknown tags hierarchy. In this case, we will convert the HTML document into a PDF document and process it as a scanned PDF where it is simultaneously processed using text and layout-based bibliographic reference extraction pipelines.

\subsection{Interfaces and Output}
In this section, we will discuss different interfaces and outputs sample our of the proposed system. Our system provides a web-based friendly interface where one interface for uploading and configuring files for processing while the other interface is responsible for displaying the results from layout-based detection. Additionally, an interface also lists all submitted processing tasks along with their output.

\subsubsection{Input interface} The input interface of our system is shown in Fig \ref{fig:inputInterface}. User can upload any file type with extension PDF, JPG, PNG, TIF, TXT, HTML and XML. In the first step, the user selects the desired file type for processing. Once the file type is selected, all available relevant pipelines are revealed. After selecting the desired processing pipeline, the user is asked to upload the desired file. Additionally, users can check an additional option on whether or not to add dummy text before the extracted text. For this purpose "\textbf{Append Dummy Text}" flag must be enabled. During the evaluation of \textit{ParsCit}\cite{CouncillGK08} we found out that appending dummy text to the start of the references text yields better results. Once all settings are done the user can trigger the processing phase by pressing \textbf{Process File} button.

\begin{figure}
    \centering
    \begin{minipage}{0.49\textwidth}
        \centering
        \includegraphics[width=0.9\textwidth]{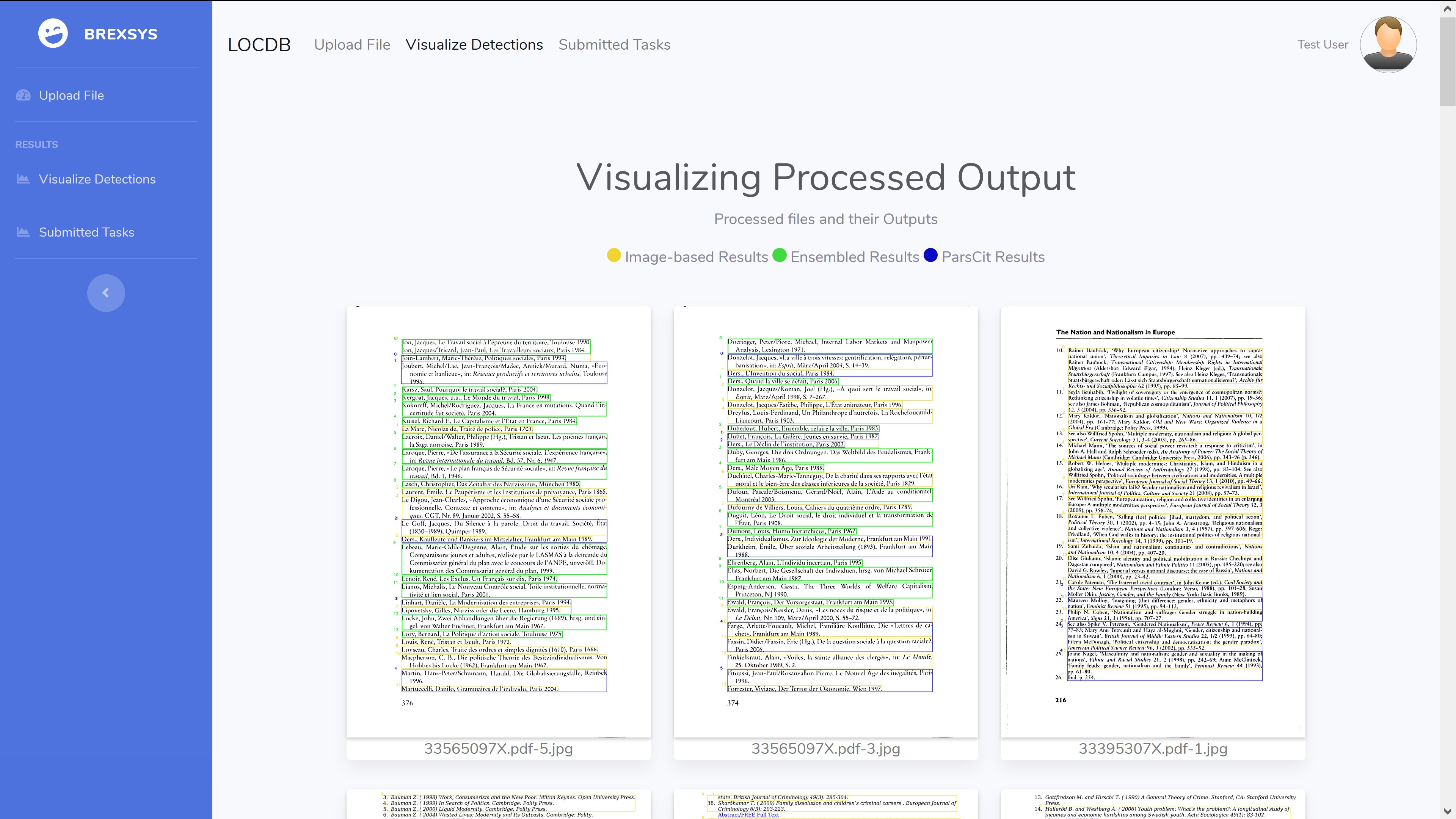} 
        \caption{Output visualizing interface}
        \label{fig:outputinterface}
    \end{minipage}\hfill
    \begin{minipage}{0.49\textwidth}
        \centering
        \includegraphics[width=0.9\textwidth]{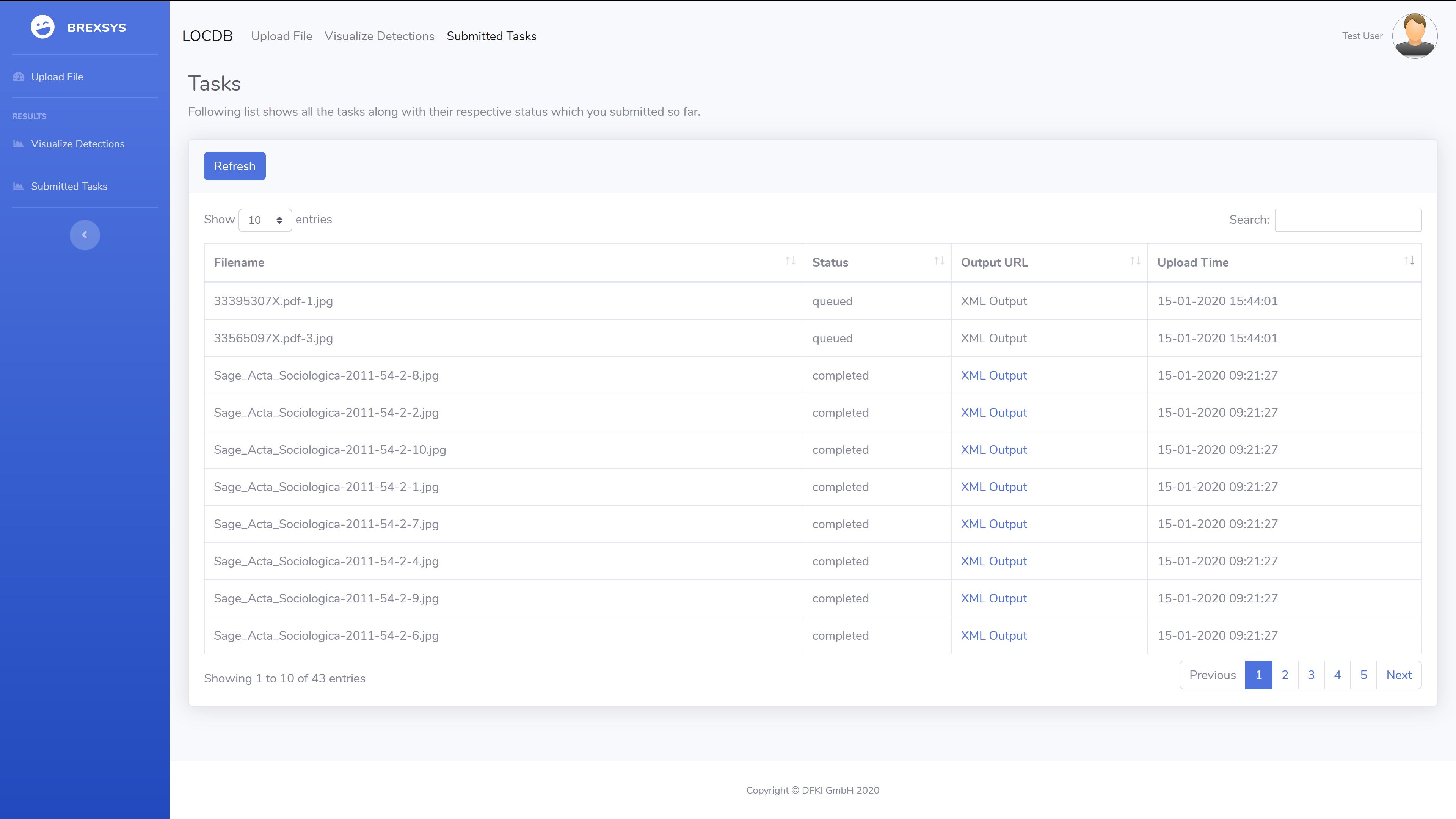} 
        \caption{Tasks status interface}
        \label{fig:tasksinterface}
    \end{minipage}
\end{figure}



\subsubsection{Output visualizing interface} Output visualizing interface is another important interface of our system where we can visualize the results of all documents processed as scanned documents. Fig \ref{fig:outputinterface} shows the output visualizing interface of our system containing all the images/scanned PDF documents already process. The detected references from both layout and text-based models are represented in different colors. Results from the layout-based model and text-based model are represented in yellow and blue respectively. While the boxes in green represent the references detected by both layout and text-based models. 

\subsubsection{Tasks status interface} This interface provides a list of all submitted processing tasks along with their history. Additionally, it also shows their current status whether a task is currently in the queue or is already processed. Fig \ref{fig:tasksinterface} shows the screenshot of the interface. Link to the XML output of each processed file is also available in front of each filename. It is to be noted that users access is protected with logins and session, therefore a user will only be able to view their processing tasks.

\subsubsection{XML Output} 
\begin{figure}
    \centering
    \begin{minipage}{0.49\textwidth}
        \centering
        \includegraphics[width=0.9\textwidth]{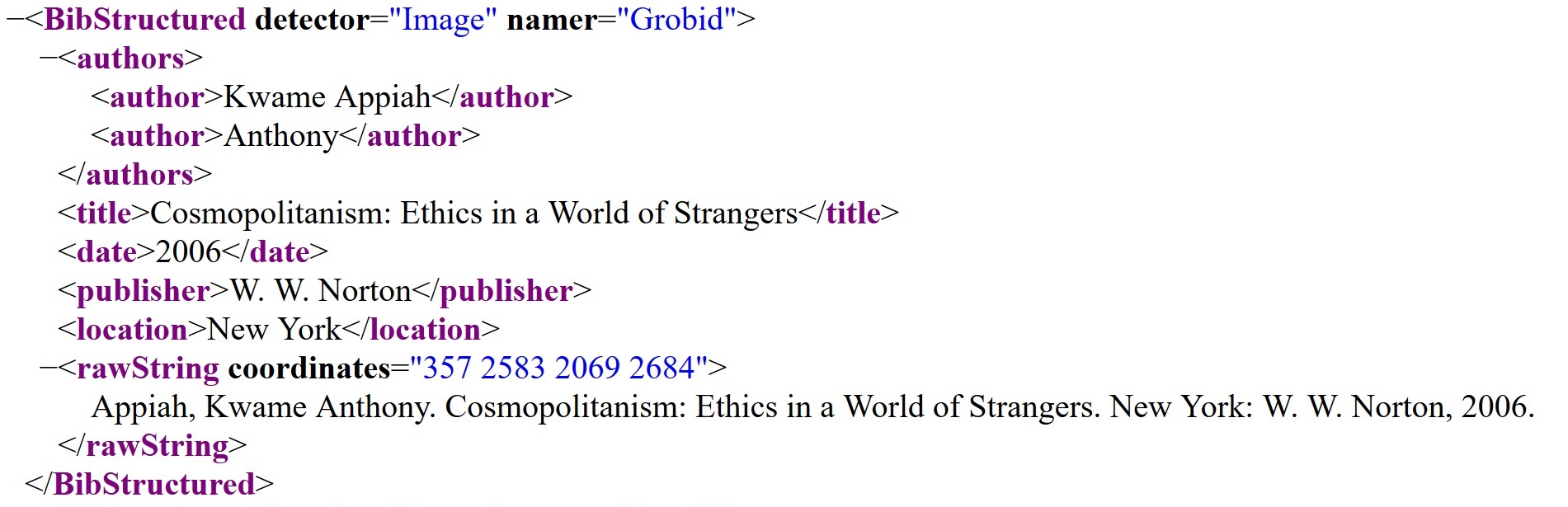} 
        \caption{XML output from layout-based model}
        \label{fig:xmlresult1}
    \end{minipage}\hfill
    \begin{minipage}{0.49\textwidth}
        \centering
        \includegraphics[width=0.9\textwidth]{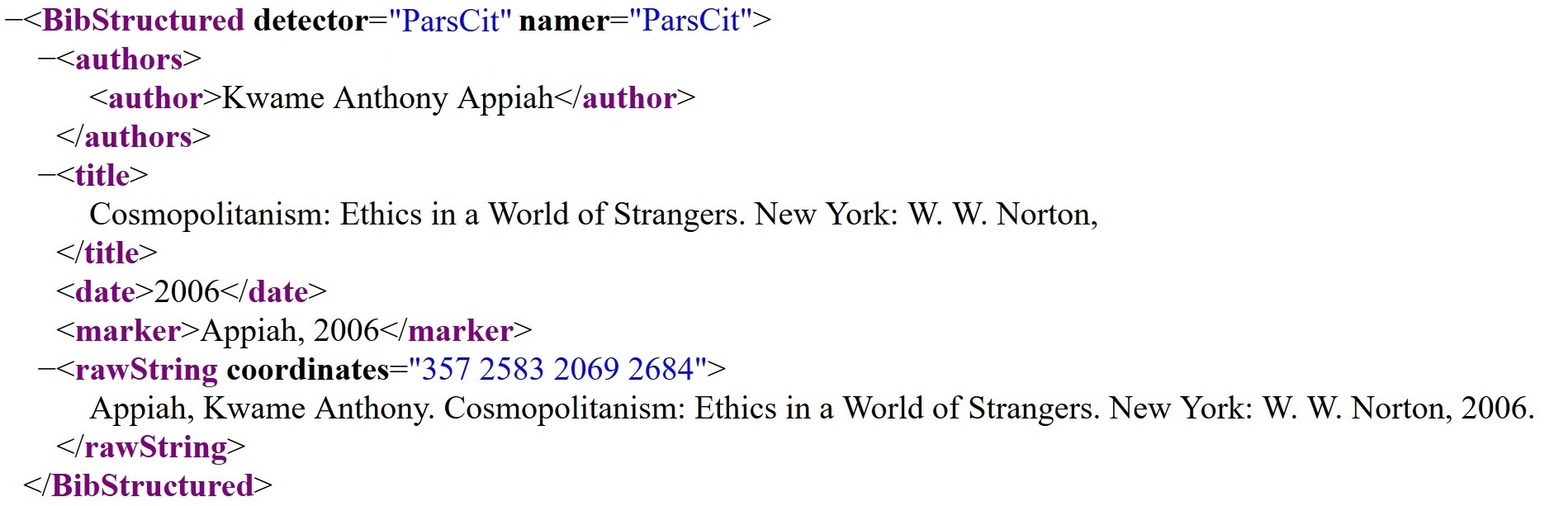} 
        \caption{XML output from Text-based model}
        \label{fig:xmlresult2}
    \end{minipage}
\end{figure}

Our system combines the output from all pipelines for the respective file type and returns it in a single XML file where results from each model are differentiated using two custom XML attributes. Fig \ref{fig:xmlresult1} and \ref{fig:xmlresult2} show output samples from layout and text-based model respectively. The custom attributes are added to identify the source of the output. First attribute is \textit{detector} which refers to the approach used to detected references i.e. image-based or \textit{ParsCit}. The second attribute is \textit{namer} which refers to the approach used to extract reference metadata i.e. author names, title, publisher, etc from raw reference string. The possible values for \textit{namer} are either \textit{ParsCit} or \textit{Grobid}.



\subsection{BRExSys Overview}
There are different pipelines in the \textit{BRExSys} framework, where the pipeline with an ensemble of Text and layout-based methods is the largest pipeline. There are several phases in ensemble pipeline. The pre-processing phase takes $\approx4.35$ seconds, followed by reference detection from image takes $\approx2.79$ seconds  which is further followed by extraction of detections takes $\approx1.95$ seconds. OCR phase takes $\approx3.63$ seconds followed by the most expensive string segmentation phase by image-based approaches which takes $\approx8.65$ seconds. Lastly, compiling both layout and text-based results in an XML file and drawing results on input image takes $\approx3.88$ and $\approx1.59$ seconds. It is worth mentioning that due to limited resources all these different services were mostly running on a single core which contributed towards using more execution time. \textit{BRExSys} was tested on a system with the following hardware specifications:
\begin{itemize}
  \item Processor: Intel® Xeon(R) E3-1245 v6 @ 3.70GHz
  \item Cores: 8
  \item Graphics: GeForce GTX 1080 Ti
  \item Memory: 64 GiB
\end{itemize}

\begin{figure*}
\centering
\subfloat[Original Document]{\frame{\includegraphics[width=0.40\linewidth]{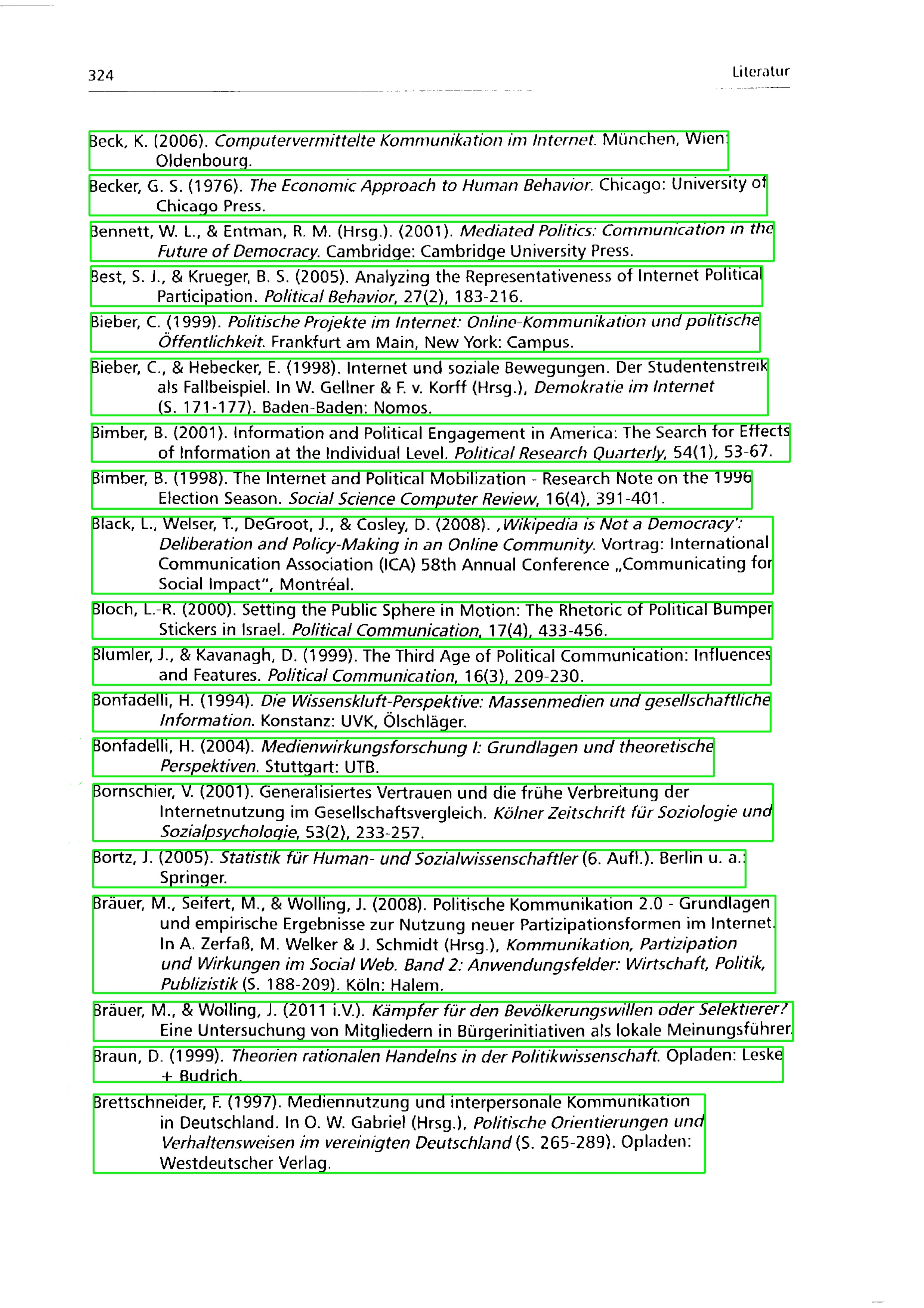}}\label{fig:original-example}}\hfil
\subfloat[Old Document Example]{\frame{\includegraphics[height=9.39cm,width=0.40\linewidth]{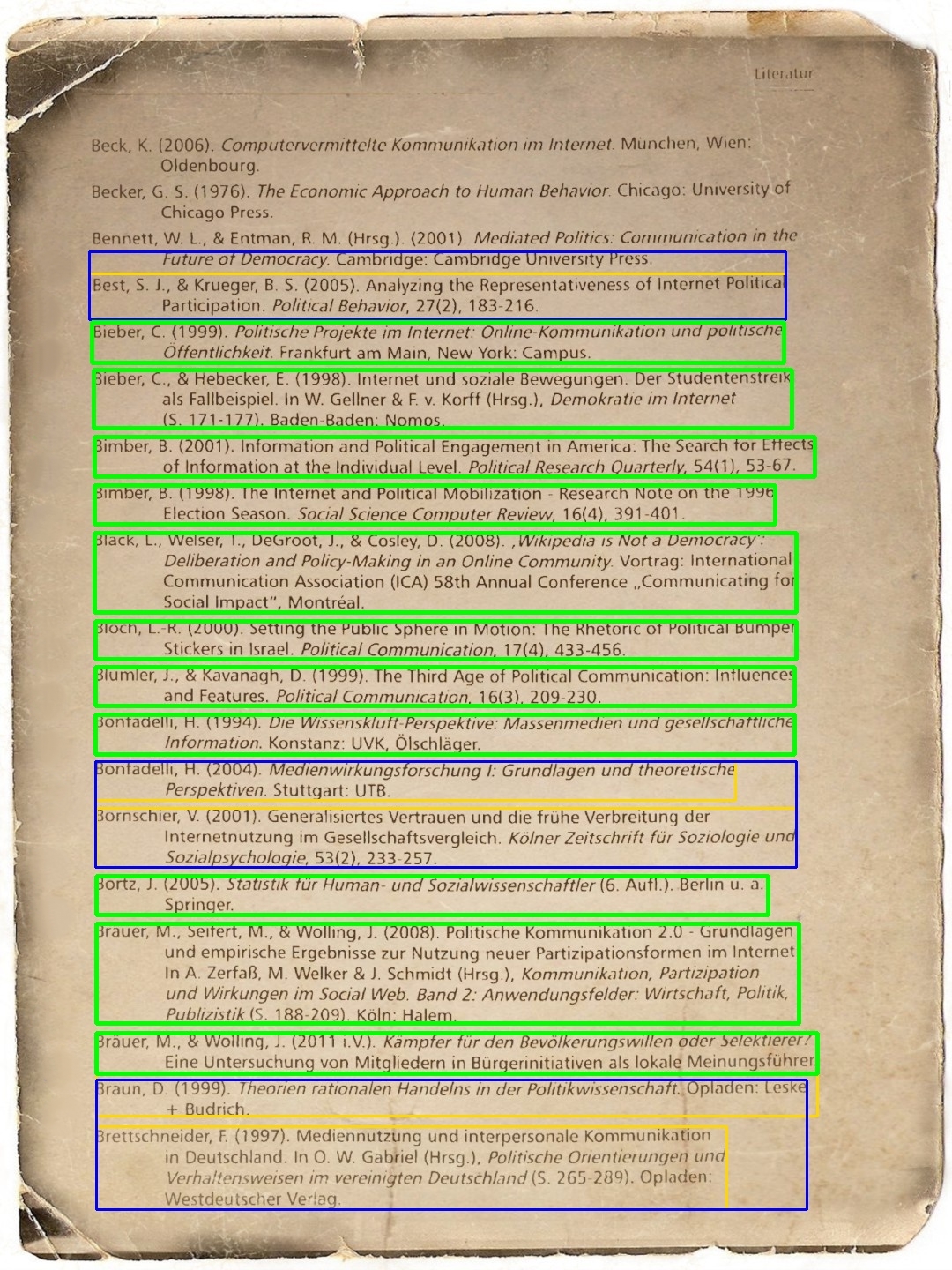}}\label{fig:torn-example}}\hfil\\
\subfloat[Dimmed Document Example]{\frame{\includegraphics[width=0.40\linewidth]{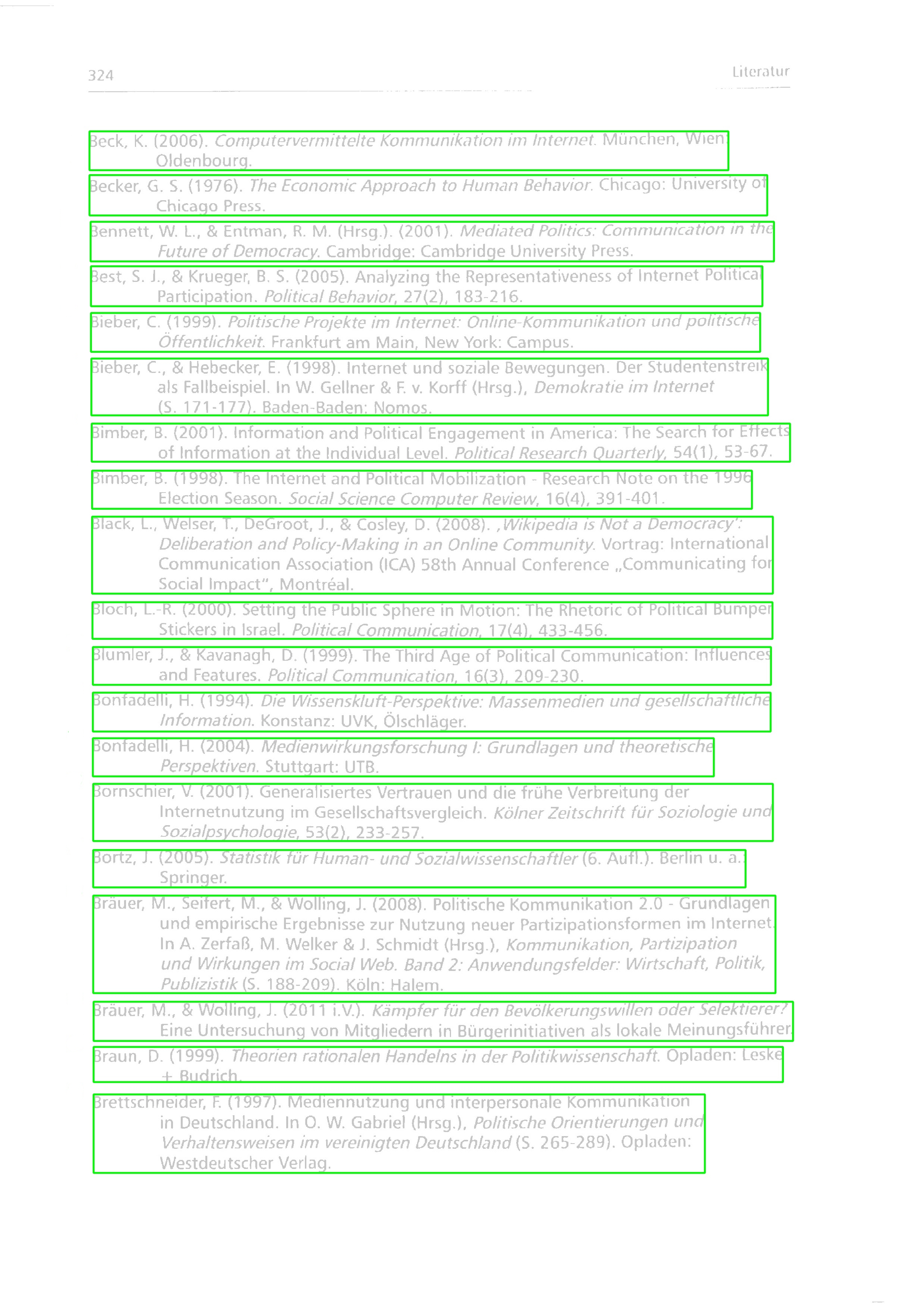}}\label{fig:dim-example}}\hfil
\subfloat[Tinted Document Example]{\frame{\includegraphics[width=0.40\linewidth]{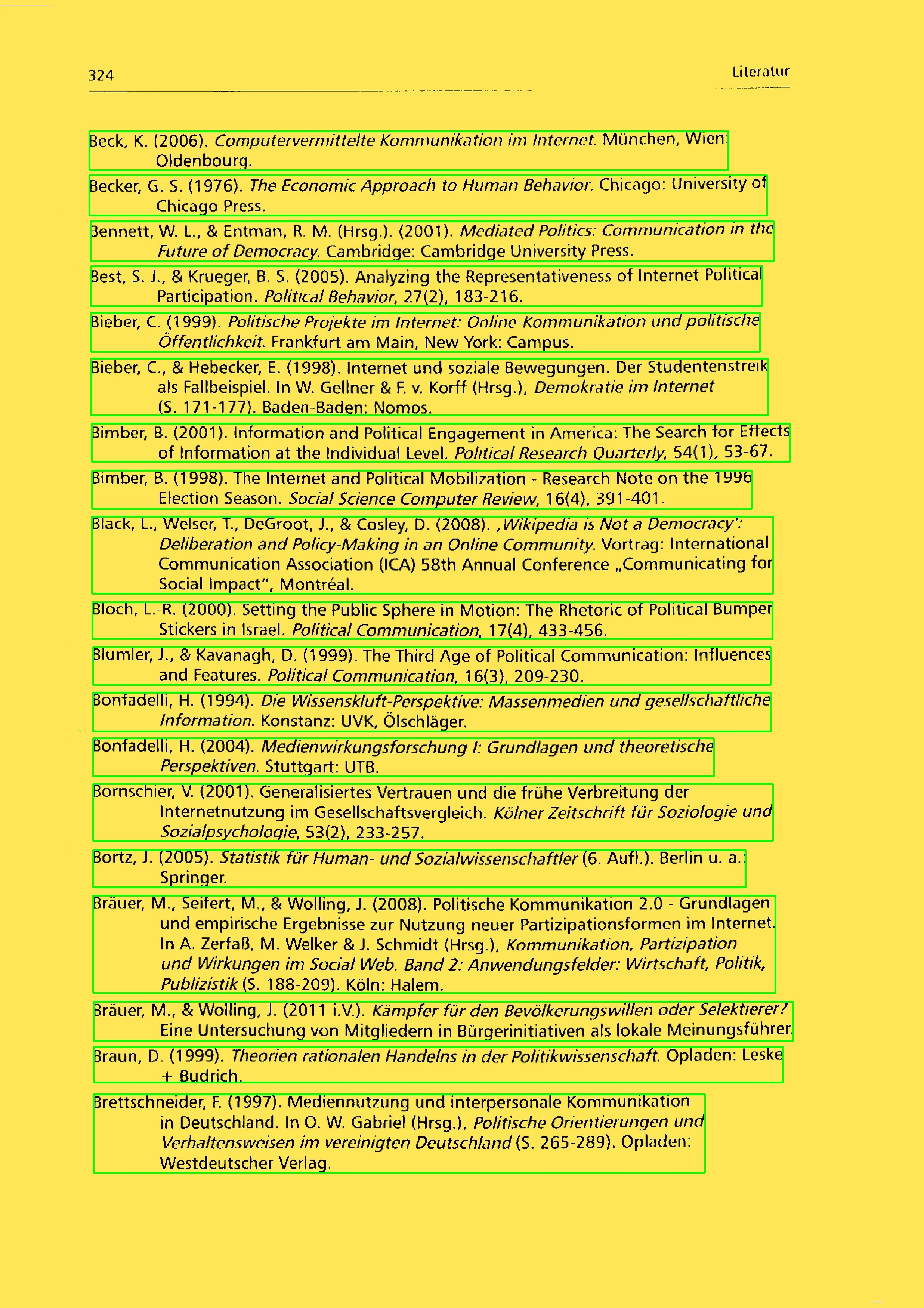}}\label{fig:tinted-example}}\hfil

\caption{Output of BRExSys for artificially created challenging hypothetical cases }
\label{fig:hypothetical_cases}
\end{figure*}

In order to evaluate \textit{BRExSys} on some more challenging cases, we prepared hypothetical examples of different cases using an actual sample from \textit{BibX} dataset. Fig.~\ref{fig:hypothetical_cases} shows output of all artificially created hypothetical cases after processing them though ensemble pipeline of layout and text-based models. Output of original image in Fig.~\ref{fig:original-example} serves as the baseline, where all references are perfectly detected. The detections of ensemble, layout-based and text-based models are represented in green, yellow and blue colors respectively.

Fig.~\ref{fig:torn-example} simulates the output of \textit{BRExSys} for an old document. The level of noise in the old document effected the output of the system, where most of references were successfully detected by the layout-based approach only missing the top three references. Similarly, text-based model also detected most of the references while missing the top three references. However, in case of text-based model, some of the references are merged and detected as one reference. Fig.~\ref{fig:dim-example} and Fig.~\ref{fig:tinted-example} simulate the example of dim and tinted document images respectively with different noise levels. However, in both cases \textit{BRExSys} successfully detected all references suggesting that only very high levels of noise may effect the out put of the system.

\section{Conclusion}
\label{conclusion}
In this paper, we presented a novel layout driven reference detection approach called "\textit{DeepBiRD}" which exploits human intuition and visual cues to effectively detect references without taking textual features into account. We also presented a dataset for image-based bibliographic reference detection which is publicly available. Later in meticulous evaluation, we pushed the boundaries of automatic reference detection and set a new state-of-the-art by a significant margin. The evaluation results suggested that the \textit{DeepBiRD} is an effective, generic, and robust approach for the problem of automatic reference detection from document images. Lastly, we presented a highly customizable framework called for automatic reference detection which employs models from different modalities i.e. image and text-based. In future work, we are aiming for improving this system to make it more robust for very noisy document scans like in Fig.~\ref{fig:torn-example}. Additionally, we intend to include the functionality where the references page is automatically detected from a scientific publication before extraction of bibliographic references.

\section*{Acknowledgment}
Authors would like to thank University Library Mannheim especially to Dr. Annette Klein, Dr. Philipp Zumstein, Laura Erhard, and Sylvia Zander for providing us dataset scans and annotations. We would also like to thank Muhammad Naseer Bajwa for his assistance during the evaluation process.

\clearpage

\clearpage
\bibliographystyle{unsrt}  


\end{document}